\documentclass[preprint,12pt]{elsarticle}

\usepackage[pagebackref,breaklinks,colorlinks,allcolors=iccvblue]{hyperref}
\usepackage[inkscapelatex=false]{svg}
\usepackage{booktabs}
\usepackage{amssymb}
\usepackage{pifont}
\usepackage{xcolor}
\usepackage{colortbl}
\usepackage{array}
\usepackage{xspace}
\usepackage{multirow}
\usepackage{adjustbox}
\usepackage{amsmath}
 \usepackage[utf8]{inputenc}
\usepackage{boldline}
\usepackage{microtype}
\usepackage{multicol}
\usepackage{multirow}
\usepackage{inconsolata}
\usepackage{svg}
\usepackage{graphicx}
\usepackage{listings}
\usepackage{epsfig,endnotes}
\usepackage[utf8]{inputenc}
\usepackage[T1]{fontenc}
\usepackage{upquote}
\definecolor{naturegreen}{RGB}{0, 156, 0}  
\definecolor{naturered}{RGB}{159, 0, 0}    

\newcommand{\up}[1]{\textcolor{naturegreen}{$\mathbf{\uparrow}$ #1}}
\newcommand{\down}[1]{\textcolor{naturered}{$\mathbf{\downarrow}$ #1}}
\newcommand{\cmark}{\ding{51}} 
\newcommand{\xmark}{\ding{55}} 

\usepackage{times}
\usepackage{latexsym}

\usepackage[T1]{fontenc}

\usepackage[utf8]{inputenc}

\usepackage{microtype}

\usepackage{inconsolata}

\usepackage{graphicx}

\begin{document}

\begin{frontmatter}
\author[uwa,unimelb]{Yihao Ding}
\author[unimelb,usyd]{Soyeon Caren Han\corref{cor1}}
\author[unimelb]{Yanbei Jiang}
\author[usyd]{Yan Li}
\author[unimelb]{Zechuan Li}
\author[unimelb]{Yifan Peng}
\cortext[cor1]{Corresponding author: caren.han@unimelb.edu.au}

\ead{yihao.ding.1@unimelb.edu.au}
\ead{caren.han@unimelb.edu.au}
\ead{lizechuan@hnu.edu.cn}
\ead{hyunsuk.chung@sydney.edu.au}

\affiliation[uwa]{
  organization={The University of Western Australia},
  addressline={35 Stirling Highway},
  city={Perth},
  postcode={6009},
  state={Western Australia},
  country={Australia}
}
\affiliation[unimelb]{
  organization={The University of Melbourne},
  addressline={Grattan Street},
  city={Melbourne},
  postcode={3010},
  state={Victoria},
  country={Australia}
}
\affiliation[usyd]{
  organization={The University of Sydney},
  addressline={Camperdown},
  city={Sydney},
  postcode={2006},
  state={New South Wales},
  country={Australia}
}
\affiliation[weillcornell]{
  organization={Weill Cornell Medicine, Cornell University},
  addressline={1300 York Avenue},
  city={New York},
  postcode={10065},
  state={New York},
  country={USA}
}



\title{SynDoc: A Hybrid Discriminative-Generative Framework for Enhancing Synthetic Domain-Adaptive Document Key Information Extraction}




\begin{abstract}
Domain-specific Visually Rich Document Understanding (VRDU) presents significant challenges due to the complexity and sensitivity of documents in fields such as medicine, finance, and material science. Existing Large (Multimodal) Language Models (LLMs/MLLMs) achieve promising results but face limitations such as hallucinations, inadequate domain adaptation, and reliance on extensive fine-tuning datasets. This paper introduces SynDoc, a novel framework that combines discriminative and generative models to address these challenges. SynDoc employs a robust synthetic data generation workflow, using structural information extraction and domain-specific query generation to produce high-quality annotations. Through adaptive instruction tuning, SynDoc improves the discriminative model's ability to extract domain-specific knowledge. At the same time, a recursive inferencing mechanism iteratively refines the output of both models for stable and accurate predictions. This framework demonstrates scalable, efficient, and precise document understanding and bridges the gap between domain-specific adaptation and general world knowledge for document key information extraction tasks. 
\end{abstract}




\end{frontmatter}



\section{Introduction}
\label{sec:intro}
With the increasing demand for domain-specific Visually Rich Document Understanding (VRDU), significant opportunities are emerging in areas such as medicine \cite{pdfvqa,mmvqa}, finance \cite{formnlu}, material science \cite{khalighinejad2024matvix}, and politics \cite{vrdu}. These areas often rely on documents that contain extensive domain-specific knowledge and sensitive information, which pose unique challenges of key information extraction (KIE) from domain specific documents. As industries increasingly turn to AI-powered solutions for document analysis, the need for robust and adaptable frameworks capable of navigating these intricacies has reached an unprecedented level.

\begin{figure}[t]
    \centering
    \tiny \includegraphics[width=\linewidth]{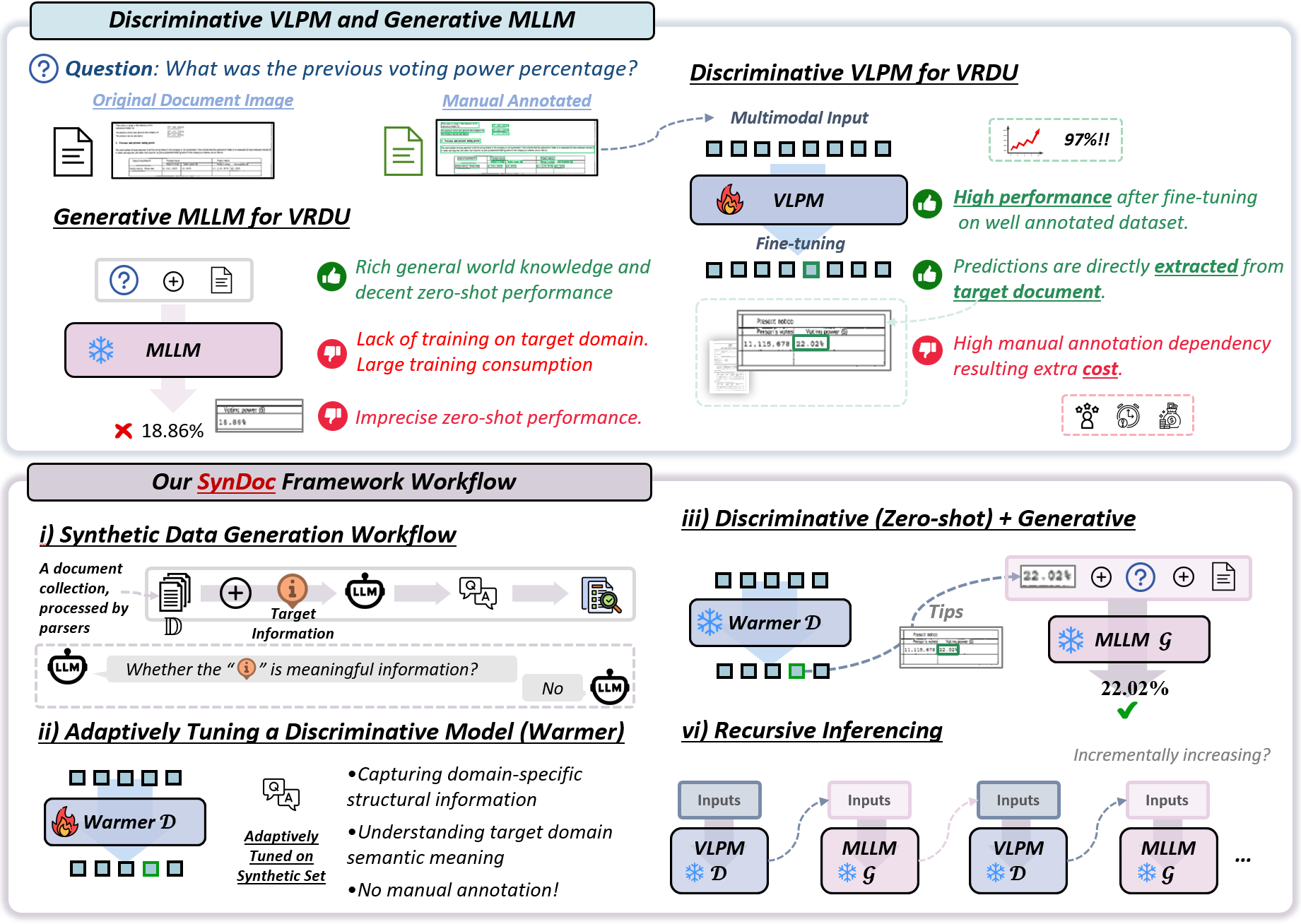}
    \caption{Comparing SynDoc with discriminative and generative document KIE frameworks.}
    \label{fig:intro}
\end{figure}

Vision-Language Pretrained Models (VLPMs) \cite{layoutlmv3,udoc,lyu2024structextv3} have demonstrated significant advancements in VRDU, normally in a \textbf{discriminative} manner by directly mapping multimodal inputs to extract key information through classification and sequence labeling. Yet, they encounter several challenges. First, they heavily depend on extensive fine-tuning datasets \cite{david}. Second, their practical use, particularly in zero-shot scenarios. Multimodal Large Language Models (MLLMs) have been applied to extract key information from VRD in a \textbf{generative} manner \cite{gpt4o,Qwen2-VL}, achieving remarkable progress due to their rich general knowledge; however, they suffer from a lack of target domain knowledge, leading to unreliable and imprecise outputs in VRDU applications. 
For instance, as shown in Figure~\ref{fig:intro}, a MLLM  extracts the \textit{present voting power} ``18.86\%" instead of the requested \textit{previous voting power}, highlighting its limitations in table structure understanding. 


Recent researches \cite{david,hu2024mplug2,feng2024docpedia,tang2024textsquare} have explored various strategies to address these challenges in document KIE, with synthetic data generation increasingly emerging as a crucial approach, driving advances in both discriminative and generative models. 
The discriminative framework uses domain-adaptive techniques in the VLPM backbone, achieving promising results through fine-tuning on curated annotated datasets \cite{david}. However, this approach remains constrained by high annotation costs and limited zero-shot performance. 
However, generative models leverage synthetic data for self-supervised pretraining \cite{hu2024mplug2,feng2024docpedia} and instruction tuning \cite{hu2024mplug,tang2024textsquare} to enhance multimodal VRD comprehension. However, the massive computational demands and suboptimal performance in zero-shot scenarios in a new domain are challenges. The synthetic generation method powered by MLLMs \cite{mmvqa} often faces issues generating meaningful or inconsistent question-answer pairs. Therefore, the field still sees a gap in research on how to improve the quality of these synthetically generated instruction-response pairs.

In this study, we propose \textbf{SynDoc}, a new hybrid framework that leverages discriminative and generative models to enhance VRDU through a multifaceted approach. 
Compared to previous studies, SynDoc offers several advantages.

First, SynDoc employs a robust synthetic data generation workflow that blends structural information extraction techniques, such as OCR (Optical Character Recognition) and PDF parsing, with multi-task inquiry generation and quality verification modules. This workflow ensures the creation of high-quality synthetic annotations that accurately reflect both document structure and content, enabling a nuanced understanding of complex domain-specific documents.  

Second, SynDoc integrates a discriminative model, referred to as the \textit{warmer}, with a generative MLLM to combine their complementary strengths. The discriminative model leverages pre-trained backbones, adaptively fine-tuned on synthetic datasets, to effectively extract domain-specific knowledge. Simultaneously, the generative model utilizes state-of-the-art MLLM to generate abstractive answers through zero-shot prompting.  

Third, SynDoc employs adaptive instruction tuning incorporating multimodal features- including text, visuals, layouts, and structural elements- with predictions from MLLMs. This approach enables the discriminative warmer to provide detailed, context-aware information, thus enhancing the outputs of the generative model. 

Finally, a key innovation in SynDoc is its recursive inferencing mechanism, where outputs from both the discriminative and generative models undergo iterative refinement through cross-feeding. This iterative process contributes to more stable and accurate responses in zero-shot settings. 

By integrating these components, we hypothesize that SynDoc offers a scalable and robust framework for domain-specific document key information extraction; we demonstrate its effectiveness on three domain-specific datasets and assess its generalizability using a cross-domain dataset.

\section{Related Work}
\paragraph{Curated and synthetic data for VRDU}
Heuristic \cite{watanabe1995layout,seki2007information} and statistical learning methods \cite{oliveira2017fast} perform well in domain-specific document understanding but rely on expert efforts, limiting cross-domain adaptability. \cite{layoutlmv3,udop,lyu2024structextv3,lilt,bros} address this limitation by employing self-supervised learning on large-scale, unannotated, and multi-source document collections such as RVL-CDIP \cite{rvlcdip}, thereby improving generalizability and multimodal comprehension in broader VRDU tasks. Fine-tuning these frameworks with curated datasets achieves state-of-the-art performance in specific VRDU tasks. However, the creation of high-quality curated datasets \cite{funsd,cord,pdfvqa} is resource-intensive, posing challenges for scalability and applicability to novel document collections. Recent research \cite{mmvqa} has explored using LLMs/MLLMs to generate synthetic datasets with well-designed prompts and human verification. Some VRDU MLLMs also create large-scale synthetic datasets to conduct self-supervised pretraining \cite{hu2024mplug2,feng2024docpedia} or instruct-tuning \cite{hu2024mplug,tang2024textsquare,zhang2024llava} to enhance multimodal document understanding. A recent work \cite{david} pretrains VRDU models with synthetic QA pairs, followed by semi-supervised refinement, achieving performance comparable to full supervision. However, there remains a limited exploration into optimizing synthetic dataset generation and integrating SoTA MLLMs for real-world applications.

\paragraph{VRDU frameworks} 
Self-supervised frameworks \cite{ernie-mmlayout,docformerv2,donut} employ diverse pretraining tasks to enhance multimodal learning, achieving strong performance on downstream tasks when fine-tuned with curated datasets. However, most discriminative models rely heavily on off-the-shelf OCR tools such as LayoutLM-series \cite{layoutlmv3,layoutxlm}, making extractive predictions vulnerable to cumulative errors from both the models and OCR systems. To mitigate this, end-to-end OCR-free frameworks \cite{donut,visfocus,lyu2024structextv3} bypass OCR dependency. Despite these advances, their smaller model sizes and limited training resources constrain world knowledge, reducing generalization without substantial annotations. LLMs/MLLMs \cite{gpt4o,gemini1.5,qwen,Idefics2}, benefiting from scaling laws, leverage extensive training to capture broad knowledge, supporting zero-shot and few-shot learning in VRD tasks \cite{icld3ie}. However, issues like hallucination and lack of domain-specific knowledge limit their reliability. Our SynDoc aims to bridge this gap by introducing an adaptively tuned discriminative warmer that provides domain-specific knowledge, which is then integrated into a generative MLLM. This approach enables the model to refine the inference process recursively, leveraging both domain-aware information and broad world knowledge to enhance accuracy and reliability.

\section{Methods}

\subsection{Overview of SynDoc}

Let $\mathbb{D}$ be a document collection within a \textit{specific domain}. We propose a framework to predict the answer to a user-provided natural language query $Q$ concerning a specific document $d \in \mathbb{D}$. This framework integrates a discriminative model $\mathcal{D}$ and a generative model $\mathcal{G}$ to address $Q$ in extractive and abstractive manners, respectively.
$\mathcal{D}$ employs pretrained backbones to capture target-domain knowledge named as a \textbf{warmer}, while $\mathcal{G}$ employs state-of-the-art LLMs/MLLMs and applies specific prompts $P$ to predict answer of in zero-shot scenarios. 

To ensure the workflow is functional, we first generate the synthetic dataset (Figure~\ref{fig:generator}).
This process begins with structural information extraction using off-the-shelf tools (e.g., OCR or PDF parsers\footnote{\url{https://github.com/PaddlePaddle/PaddleOCR} or \url{https://pypi.org/project/pdfminer/}}). 
Next, synthetic domain-specific queries are generated using MLLMs. 
Therefore, $\mathcal{D}$ incorporates multimodal representations, including textual, visual, layout, and structural features, along with predictions from MLLM.
During inference, the outputs from $\mathcal{D}$ and $\mathcal{G}$ undergo iterative refinement through cross-feeding until they achieve convergence (e.g., stable predictions). 
The following subsections describe four key modules in SynDoc: Synthetic Data Generator, Discriminative Warmer Architecture, Adaptive Instruction Tuning, and Recursive Inference.

\subsection{Synthetic Data Generator}

\paragraph{VRD Structure Parsing}
\begin{figure*}[t]
    \centering
   \includegraphics[width=\linewidth]{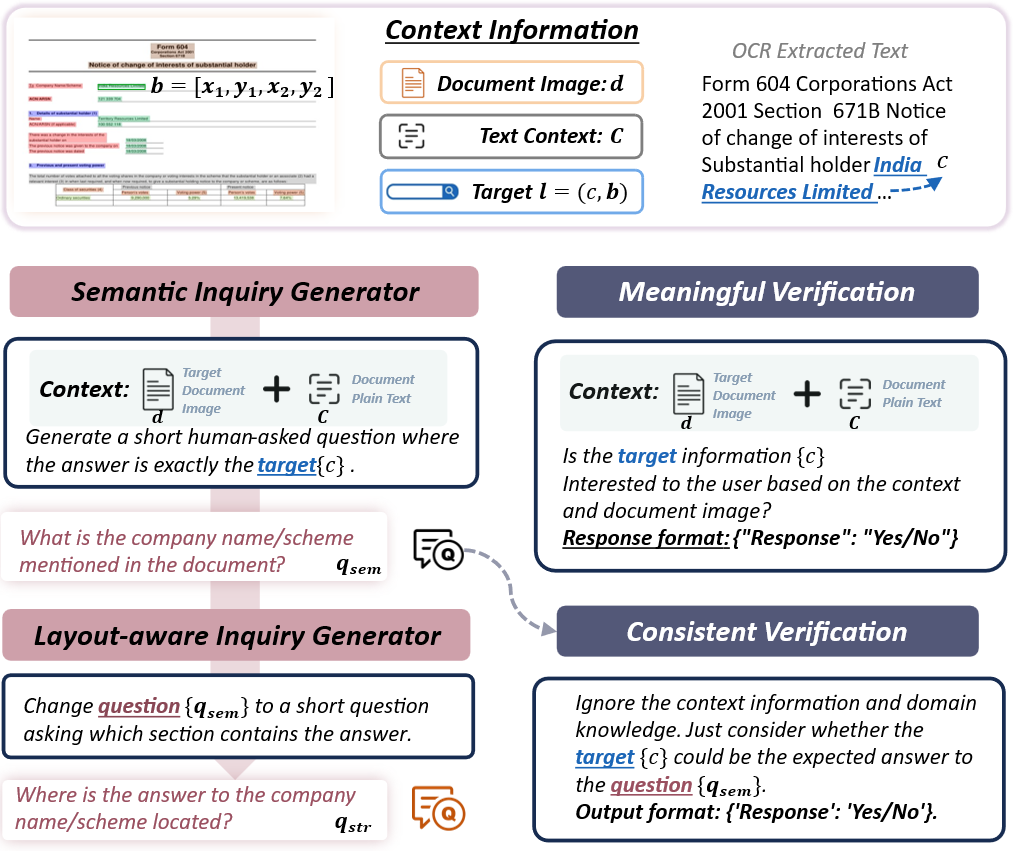}
    \caption{Workflow of the Synthetic Data Generator.}
    \label{fig:generator}
\end{figure*}  
We use off-the-shelf tools to extract the text content and layout structure of a target document collection (Figure~\ref{fig:generator}). 
For document images, we employ vision-based OCR tools to get text line entities $L$. Each $l =(b, c)\in L$ contains the bounding boxes $b$  with corresponding textual content $c$.
We use ($x_{min}, y_{min}, x_{max}, y_{max}$) to represent coordinates of each box.
For text-embedded PDF files, we employ the PDF parsing tools to acquire text line or document semantic entity sets $L$ (e.g., paragraph, list, section) along with more accurate structural information. 


\paragraph{MLLM-driven Inquiry Generation} 

For $\mathcal{D}$ to capture knowledge from the target domain, we propose a MLLM-driven workflow with two modules (Figure~\ref{fig:generator}). 
\textit{i) Multi-Task Inquiry Generation} produces diverse inquiries to instruct-tune $\mathcal{D}$ to enhance its structural and semantic understanding of the domain. 
Specifically, a set of text lines is randomly selected and fed to an LLM to generate two types of QA pairs. First, \textit{Semantic} QA pairs guide $\mathcal{D}$ to extract target information from a document. By inputting the target entity content along with its document and context information into an MLLM, we generate pairs $(q_{sem}, c)$, where $c$ is the answer to the generated question $q_{sem}$. Second, \textit{Spatial-aware} QA pairs facilitate $\mathcal{D}$ in capturing both semantic and spatial correlations. Here, we transform $q_{sem}$ into $q_{spt}$ by identifying the document region (e.g., top-left, top-middle, top-right) where the target information $c$ is located. 
\textit{ii) Multi-Aspect Quality Verification} is implemented to filter out low-quality questions by assessing factors including meaningfulness and question-answer consistency. 
It first determines whether $c$ is relevant to the end user (e.g., ``Is the target {information} interesting to the end user?").
It then verifies that $c$ adequately answers $q_{sem}$ (e.g., ``Whether the target information {$c$} could be expected answer of a question {$q_{sem}$}?'').
\begin{figure}[th]
    \centering
 \includegraphics[width=0.95\linewidth]{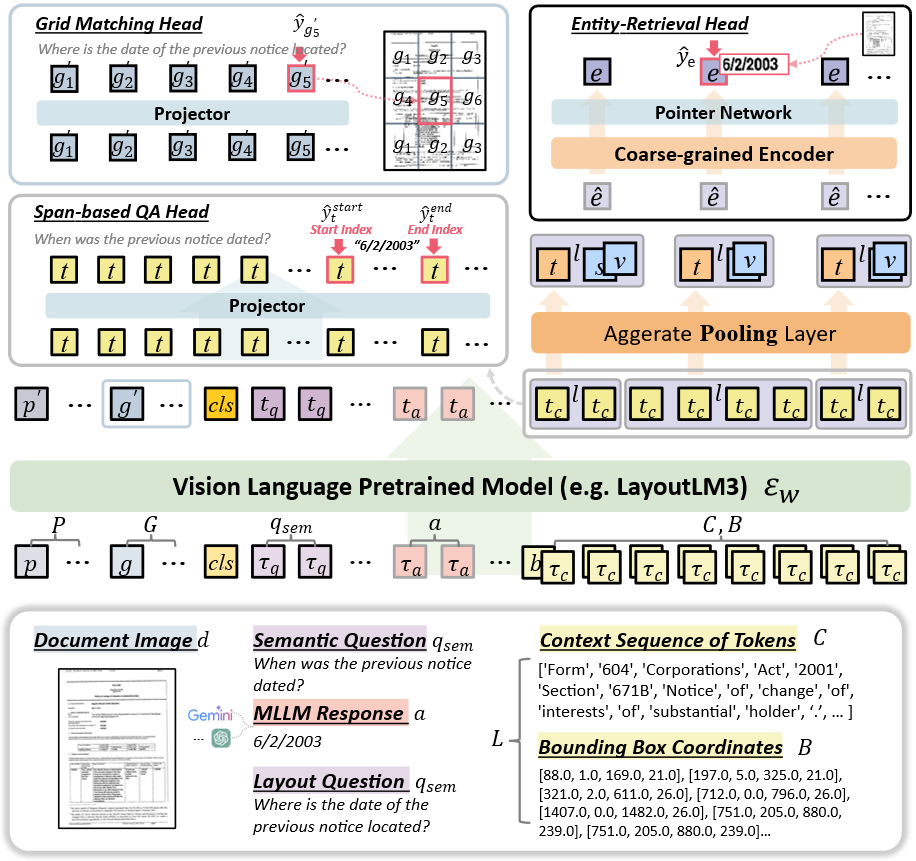}
    \caption{Architecture of the discriminative Warmer}
    \label{fig:warmer_architecture}
\end{figure}  
\subsection{Warmer Architecture}

\textbf{Warmer} ($\mathcal{D}$) utilizes a vision-language pre-trained model (VLPM) as its backbone, optimized for discriminative answer extraction through adaptively tuning on synthetic datasets. The adopted VLPM is pre-trained on layout-aware tasks and fine-tuned on well-annotated datasets, exhibiting decent performance in targeted VRDU tasks. To address zero-shot scenarios, we design the warmer architecture based on the VLPM backbone, enabling $\mathcal{D}$ to learn multi-aspect domain-aware knowledge from synthetic datasets. We will first introduce the initial feature representation of $\mathcal{D}$ and then describe the detailed architecture.

\paragraph{Initial Feature Representation}
For a synthetically acquired entity set $L$ of document $I_d$, a pretrained vision model extracts visual representation $v$ from $b$ and a text model extracts sentence representation $s$ from $c$ \cite{mmvqa}. $b$'s coordinates are linearly projected to match $s$ \cite{lxmert}.  
A textual sequence $C = \{\tau_{i}\}_{i=1}^n$ encodes context, summed with projected coordinates $B = \{b_i\}_{i=1}^n$ and, if relevant, concatenated with document image patches $P$. For each semantic query $q_{sem}$, the MLLM-generated answer $a$ can aid localization.  
Grid embeddings $G = \{g_i\}_{i=1}^{j \times k}$ result from resizing and flattening pixel data over a $j \times k$ grid of the document image.

\paragraph{Detailed Architecture.}  
$\mathcal{D}$ processes the input word sequence ($q$, $a$, $C$, $P$ and $B$). These inputs are passed through a VLPM backbone, $\mathcal{E}_{w}$, to derive embedded feature representations:  
\begin{equation}
(P', G', T_q, T_a, T_c) = \mathcal{E}_{w}(P, G, q, a, C+B)
\end{equation}
where $T$ represented corresponding encoded textual features, while $P'$ and $G'$ represent the encoded patch and grid features, respectively.  

For each $l \in L$ extracted using parsing tools, a pooling layer aggregates the token features to obtain the entity-level representation $e$. 
\begin{align}
\hat{e} &= \text{Pooling} \left( \{ \mathcal{E}_{w}(c_{i}), c_{i} \in c \} \right)\\
e &= \hat{e} \oplus v \oplus s
\end{align}
%
The enhanced entity features, $E = \{e_l\mid l\in L\}$, are processed by an \textbf{Entity-Retrieval Head}, which includes a coarse-grained transformer encoder for improving entity-level contextual understanding and a pointer network \cite{mmvqa} to predict the final entity index.  
Additionally, a fine-grained \textbf{Span-based QA Head} is employed to predict the start and end indices of the answer span based on the input query $q$. A \textbf{Grid Matching Head} is introduced to enhance structural understanding within the target domain. This matching head predicts the grid index of the input set $G'$ by leveraging specially aware queries. A different head is trained on diverse stages to enable warmer capture of adequate domain-specific knowledge.

\subsection{Adaptively Warmer Tuning}
Step-by-step training enables the warmer $\mathcal{D}$ to effectively adapt to the target domain, starting with \textbf{structural adaptation} to enhance the domain-specific structural understanding, followed by the task-oriented \textbf{semantic adaptation} for locating target information based on the input query. 

\textbf{Structural Adaptation} enhances both semantic and layout understanding by guiding $\mathcal{D}$ to identify the most relevant document grid $g' \in G'$ for a given structural query $q_{str}$. 
For example, given the query ``\textit{Where is the date of the previous notice located?}'', $\mathcal{D}$ predicts the grid $g_5$ that contains the answer (Figure~\ref{fig:warmer_architecture}). A pointer network computes the logit for each candidate grid \cite{mmvqa}, and the probability over grids is obtained using the softmax function. The model optimization employs the cross-entropy loss function to compute the structure adaptation loss \( \mathcal{L}_{str} \):
\begin{equation}
\mathcal{L}_{str} = - \sum\nolimits_{g' \in G'} y_{g'} \log \hat{y}_{g'}
\end{equation}
where \( y_{g'} \) represents the ground truth indicator of each grid. This adaptation process ensures that the model effectively learns to associate structural queries with relevant document regions, improving both retrieval accuracy and layout-aware reasoning.

\textbf{Semantic Adaptation} enables $\mathcal{D}$ to pretrain on a synthetic semantic QA set $P$, allowing it to better understand document image $I_d$ and $q_{sem}$ for zero-shot extractive QA in real-world scenarios. The model employs two extractive QA heads: a fine-grained, span-based QA head and a coarse-grained entity-retrieving head. The fine-grained head predicts the start and end token indices using a linear projector, with the cross-entropy loss defined as:
\begin{equation}
\small
\mathcal{L}_{fg} = - \sum\nolimits_{t \in \mathcal{E}_{w}(c)} y_t^{\text{start}} \log \hat{y}_t^{\text{start}} + y_t^{\text{end}} \log \hat{y}_t^{\text{end}}
\end{equation}
where $y_t^{\text{start}}$ and $y_t^{\text{end}}$ denote the ground truth indices, while $\hat{y}_t^{\text{start}}$ and $\hat{y}_t^{\text{end}}$ represent the predicted probabilities after Softmax. 

The coarse-grained entity retrieving head retrieves entities based on entity logits and is optimized with a cross-entropy loss function:
\begin{equation}
\mathcal{L}_{cg} = - \sum\nolimits_{e \in E} y_e \log \hat{y}_e
\end{equation}
where $y_e$ represents the ground truth probability distribution over the entity set $E$, and $\hat{y}_e$ is the predicted Softmax normalised probability. The final optimization objective combines both losses as:
\begin{equation}
\mathcal{L} = \lambda_{fg} \mathcal{L}_{fg} + \lambda_{cg} \mathcal{L}_{cg}
\end{equation}
where $\lambda_{fg}$ and $\lambda_{cg}$ control the balance between the fine-grained and coarse-grained QA losses. During the semantic adaptation process, different synthetic subsets may be selected based on \textit{Multi-Aspect Quality Verification} results, possibly leading to varying performance, as described in Section~\ref{sec:warmer_performance}.

\subsection{Recursively Inferencing}

\begin{figure}[t]
    \centering
    \tiny \includegraphics[width=\linewidth]{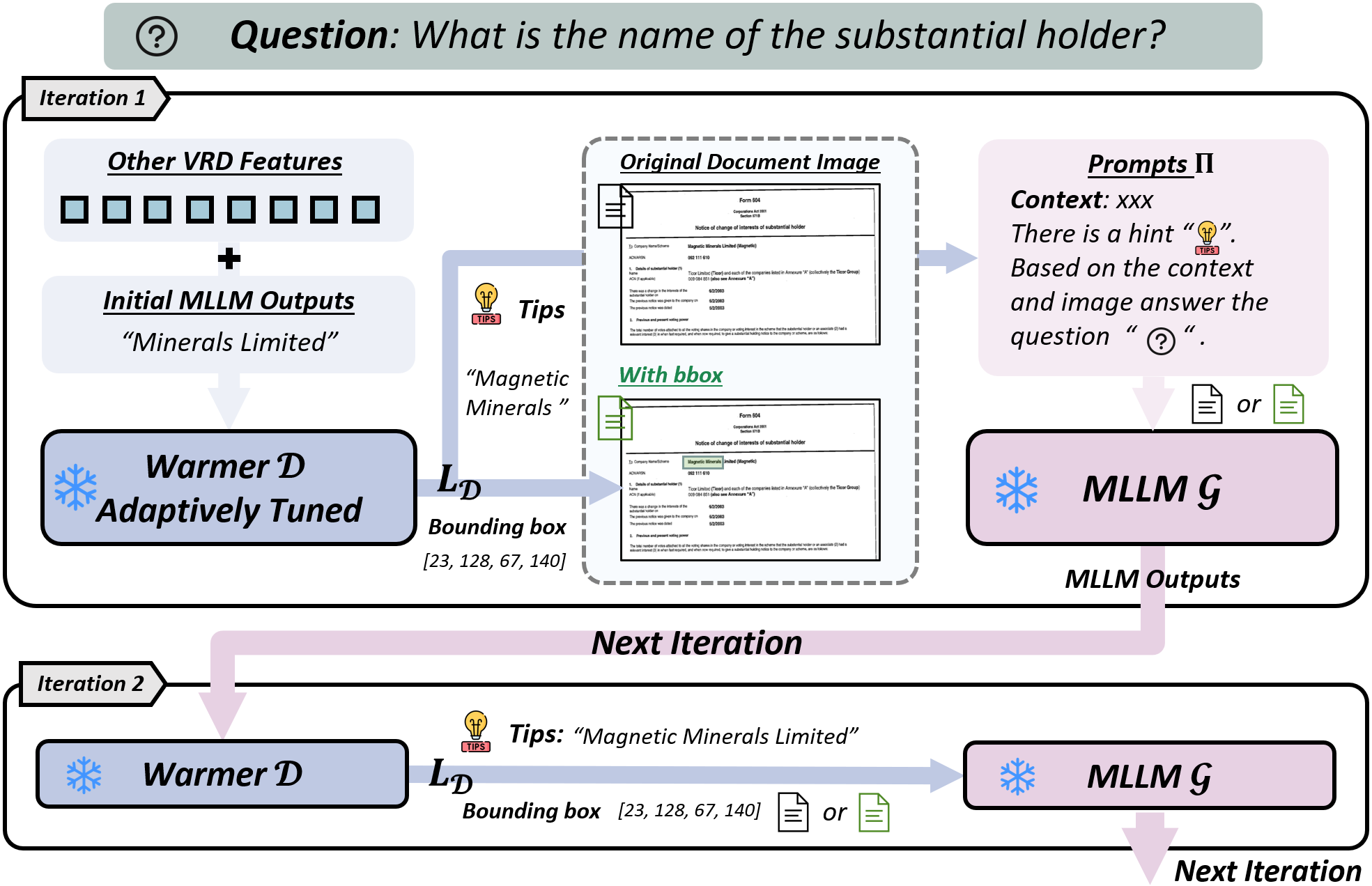}
    \caption{An illustration of the recursively inferencing framework for zero-shot question answering on VRDs. Given a question, \textit{``What is the name of the substantial holder?''}, the initial MLLM output is enhanced using retrieved entity hints (\(L_{\mathcal{D}}\)) from the adaptively tuned warmer. Bounding box hints and other VRD features guide MLLM toward more precise answers in subsequent iterations.}
    \label{fig:recursive_inference}
\end{figure}

We propose a recursively inferencing framework to harness $\mathcal{D}$ and $\mathcal{G}$ for zero-shot question answering on VRDs  (Figure~\ref{fig:recursive_inference}). The retrieved top-$k$ entities  $L_{\mathcal{D}}$ serve as domain-specific guidance to enhance MLLM responses. 
%
Originally, given the prompt $(I_d, C, q_{sem})  \rightarrow \Pi $, $\mathcal{G}$ generates an answer $A_{\mathcal{G}}$. 
In the $t$-th recursive process, $\mathcal{D}$ refines its retrieval based on the previous $A_{\mathcal{G}}^{(t)}$, leading to an updated prompt that integrates the extracted entity information:
\begin{align}
L_{\mathcal{D}}^{(t+1)} &= \mathcal{D}(A_{\mathcal{G}}^{(t)})\\
\Pi^{(t+1)} &= \text{UpdatePrompt}(\Pi^{(t)}, L_{\mathcal{D}}^{(t+1)})\\
A_{\mathcal{G}}^{(t+1)} &= \mathcal{G}(\Pi^{(t+1)})
\end{align}
This allows $\mathcal{G}$ to acquire more domain-specific knowledge, improving its ability to comprehend and locate question-relevant information within the context with greater accuracy and reliability.
The iterative refinement process enhances both extractive and generative responses over time.

\section{Experimental Settings}
\subsection{Datasets}
We used four document key information extraction datasets from different domains to evaluate SynDoc: FormNLU (financial forms) \cite{formnlu}, CORD  (receipts) \cite{cord}, Ephoie (exam papers) \cite{vies}, and FUNSD \cite{funsd} (multi-domains). (Appendix~A.1 for more details). 
Form-NLU was further divided into Printed (F-P) and Handwritten (F-H) subsets.
The document images in each \textbf{test set} were processed using the \textit{Synthetic Data Generation} module to produce synthetic structure annotations and QA pairs with verification results. During inference, QA pairs or key-value/question pairs from the original dataset are utilized.

For the FUNSD and CORD datasets, we utilized the processed test sets from \cite{luo2024layoutllm}. For Form-NLU and Ephoie, we converted the key-value pairs into QA pairs for inference. 
Consistent with \cite{docvqa,luo2024layoutllm}, we used the Averaged Normalized Levenshtein Similarity (ANLS) as our primary \textbf{evaluation metric}.

\subsection{Baselines and Implementation Details}

We evaluated SynDoc against a range of SoTA baselines, including open-source models (Qwen2-VL~\cite{Qwen2-VL}, Idefics2~\cite{Idefics2}, InternVL2~\cite{internvl}) and proprietary systems (GPT-4o~\cite{gpt4o}, Gemini 1.5~\cite{gemini1.5}). These models were selected for their demonstrated effectiveness on document-centric tasks, including comprehension, retrieval, and question answering. For consistency, all MLLMs were assessed under default inference settings in the HuggingFace environment\footnote{\url{https://huggingface.co/}} using up to $2\times$ A100 80G GPUs. Warmer tuning employed a batch size of 16, a learning rate of $2\mathrm{e}{-5}$, and the AdamW optimizer; Table~\ref{tab:warmer_tuning} summarizes performance across different tuning epochs.

\section{Results and Discussion}
\subsection{Main Results}
\label{sec:main_results}
\begin{table}[t]
    \centering
    \small
    \renewcommand{\arraystretch}{1.1} 
    \setlength{\tabcolsep}{8.5pt} 
    \resizebox{0.75\linewidth}{!}{%
    \begin{tabular}{l c c c c | c}
        \hline
        \bf Model & \bf F-P & \bf F-H & \bf CORD & \bf Ephoie & \bf FUNSD \\
        \hline
        \bf Idefics2 & 57.54 & 33.31 & 54.45 & 15.22 & 62.11\\
        \bf InternVL2 & 66.56 & 45.47 & 66.84 & 68.92 & 74.95\\
        \bf Qwen2-VL & \underline{78.05} & 43.65 & 77.86 & 70.36 & 79.12\\
        \bf GPT-4o & 76.16 & 56.49 & 79.05 & 79.40 & 80.05\\
        \bf Gemini & 76.09 & \underline{66.86} & \underline{84.35} & \underline{81.82} & \underline{83.56}\\
        \hline \hline
        \multicolumn{6}{l}{\bf SynDoc (Gemini)}\\ \hline
        \bf Top-1 & \cellcolor{green!20}80.29 & \cellcolor{green!20}67.73 & \cellcolor{green!20}85.19 & 81.80 & 82.77\\
        \bf Top-$K$ & \cellcolor{green!20}81.60 & \cellcolor{green!20}66.90 & 83.57 & 81.33 & 82.12\\
        \bf Top-1 R & \cellcolor{green!20}80.29 & \cellcolor{green!20}67.73 & \cellcolor{green!20}85.19 & \cellcolor{green!40}\textbf{82.15} & 83.02\\
        \bf Top-$K$ R & \cellcolor{green!40}\textbf{81.91} & \cellcolor{green!20}68.09 & \cellcolor{green!20}84.57 & 81.58 & 82.40\\
        \hline
        \bf w/bbox & \cellcolor{green!20}80.93 & \cellcolor{green!40}\textbf{68.13} & \cellcolor{green!40}\textbf{85.40} & \cellcolor{green!20}82.08 & \cellcolor{green!40}\textbf{83.87}\\
        \hline
    \end{tabular}%
    }
    \caption{Results using Zero-shot MLLM. The last row shows the best configuration with bounding boxes.} 
    \label{tab:main_results}
\end{table}

Table~\ref{tab:main_results} shows that proprietary models generally outperform their open-source counterparts. This advantage is particularly evident in complex scenarios (e.g., F-H and Ephoie). 
Among similarly sized open-source MLLMs, Qwen2-VL achieves the highest performance, benefiting from its extensive multimodal training data and advanced OCR capabilities. Intern-VL2 also demonstrates strong performance across all datasets, whereas Idefics2 encounters challenges, particularly with structurally complex documents in Ephoie. 

Since Gemini shows better performance across most benchmark datasets compared to GPT-4o, we present the results of the Gemini-based SynDoc framework. Overall, incorporating adaptively tuned warmer knowledge into MLLMs enhances performance on domain-specific datasets; however, it may introduce noise in cross-domain benchmarks such as FUNSD. The results also suggest that employing top-$K$ candidate hints or recursive inference (top-$K$ R) substantially improves MLLM performance in zero-shot scenarios. 

\subsection{Warmer Performance Analysis}
\label{sec:warmer_performance}

Here, we evaluated the effectiveness of the \textit{Synthetic Data Generation} workflow and \textit{Warmer}’s capability to capture domain-specific knowledge in three aspects: adaptive tuning strategies, top-$k$ entity retrieval variations, and comparisons of pretrained backbones. These experiments aim to determine whether the proposed methods optimize Warmer’s performance and enhance information extraction in a zero-shot setting, ultimately providing strong support for downstream MLLM inference. Notably, without tuning on synthetically generated data, as shown in Table~\ref{tab:warmer_tuning}, the pretrained Warmer failed sto retrieve any meaningful information from the target document context.

\begin{table}[t]
    \centering
    \small
    \resizebox{0.8\linewidth}{!}{%
    \begin{tabular}{@{}ccccccccc@{}}
        \toprule
        \textbf{Adapt} & \textbf{St} & \textbf{Prior} & \textbf{F-P} & \textbf{F-H} & \textbf{CORD} & \textbf{Ephoie} & \textbf{FUNSD} \\
         \midrule
        N/A & \xmark & \xmark & 0 & 0 & 0 & 0 & 0 \\
        \midrule
        1 & \xmark & \xmark & 31.39 & \underline{18.18} & 41.48 & 19.23 & 44.37 \\
        2 & \xmark & \xmark & 42.56 & 16.41 & 46.71 & 20.64 & \underline{48.66} \\
        3 & \xmark & \xmark & 33.87 & 14.61 & 41.16 & 22.74 & 42.77 \\
        4 & \xmark & \xmark & \underline{44.23} & 12.23 & \underline{50.44} & \underline{23.78} & 44.67 \\
        \midrule
        1 & \xmark & \cmark & 59.26 & 30.67 & 65.6 & 22.94 & 56.83 \\
        2 & \xmark & \cmark & 65.67 & \underline{31.63} & \underline{66.37} & 22.06 & 57.77 \\
        3 & \xmark & \cmark & 64.68 & 27.85 & 65.9 & \underline{25.48} & 57.43 \\
        4 & \xmark & \cmark & \underline{65.75} & 29.31 & 65.08 & 24.76 & \underline{59.86} \\
        \midrule
        1 & \cmark & \cmark & 62.67 & 30.25 & 66.21 & 24.12 & 58.08 \\
        2 & \cmark & \cmark & 66.03 & \textbf{31.64} & \textbf{67.26} & 24.13 & 58.05 \\
        3 & \cmark & \cmark & 65.2 & 28.83 & 63.94 & 25.29 & 61.01 \\
        4 & \cmark & \cmark & \textbf{66.19} & 28.29 & 66.25 & \textbf{27.16} & \textbf{61.24} \\
        \bottomrule
    \end{tabular}}
    \caption{Results under various Warmer Adaptive Tuning  Configurations. Adapt - Four types of adaptive tuning sets: (1) full synthetic set, (2) meaningful verification filtered set, (3) consistency verification filtered set, and (4) dual verification filtered set. St - structure adaptation. Prior - prior MLLM outputs.}
    \label{tab:warmer_tuning}
\end{table}

\paragraph{Adaptive Tuning Strategies} 
We first evaluated the adaptive tuning methods in adaptive tuning sets, prior MLLM outputs, and structural-adaptive tuning.

\textit{i) Effects of adaptive tuning sets.} Table~\ref{tab:warmer_tuning} shows that both verification methods improve performance and enhance domain adaptation. However, meaningfulness verification consistently provides performance gains, while consistency verification can sometimes negatively affect tuning. This negative impact may be attributed to OCR errors, which can lead to inaccurate MLLM justifications.

\textit{ii) Impact of prior MLLM outputs.} Table~\ref{tab:warmer_tuning} also shows that incorporating MLLM outputs as Warmer input helps Warmer efficiently locate relevant information with improved accuracy. 
For instance, ANLS performance improves markedly from 41.48 to 66.37 on CORD, 44.37 to 61.24 on FUNSD, and 19.23 to 27.16 on Ephoie.
Training Warmer on the synthetic dataset enables it to extract more relevant content, with the MLLM output serving only as a reference. Providing MLLM inputs helps Warmer efficiently locate target answers, but its knowledge is primarily derived from the synthetic dataset rather than being solely influenced by MLLM outputs.

\textit{iii) Structural Adaption Tuning (St)} is introduced to enhance the Warmer model by improving its comprehension of layout and semantic correlations within a specific domain. 
Table~\ref{tab:warmer_tuning} consistently demonstrates its efficacy across all datasets. 
The performance improvements are particularly notable, with CORD increasing from 50.44 to 67.26, FUNSD from 44.67 to 61.24, and Ephoie from 22.74 to 27.16, when comparing the best results with and without St.-A. 
The result indicates that the proposed self-supervised structural adaptation effectively warms up the Warmer, enabling it to capture richer structural and semantic correlations while enhancing subsequent semantic adaptation.

\begin{figure}[t]
    \centering
    \tiny \includegraphics[width=\linewidth]{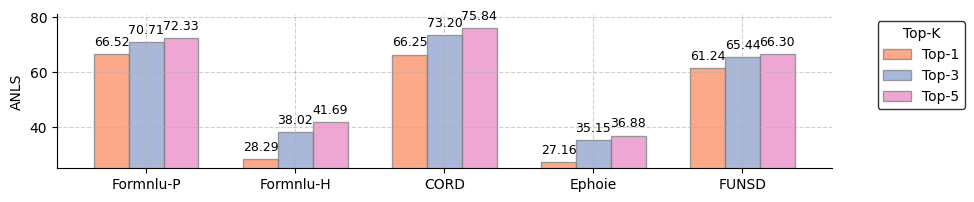}
    \caption{Top-$K$ retrieved entity performance using LayoutLMv3 as the backbone.}
    \label{fig:warmer_topk}
\end{figure}

\paragraph{Top-$K$ Retrieved Entity Performance}
While the maximum likelihood prediction of warmer may not always yield the most relevant result, leveraging Top-$K$ likelihood predictions can enhance the MLLM’s ability to locate the correct answer more effectively, as demonstrated in Section~\ref{sec:main_results}. We compare the Top-1, Top-3, and Top-5 retrieved entities, selecting the highest ANLS entity when multiple entities are given. As illustrated in Figure~\ref{fig:warmer_topk}, the Top-3 predictions significantly improve the retrieval of relevant information compared to Top-1. However, the performance gain between Top-3 and Top-5 is marginal. Notably, the improvement from Top-1 to Top-3 is more pronounced for datasets with lower OCR accuracy, indicating the benefit of broader retrieval in error-prone scenarios.

\paragraph{Various Warmer Backbones.} 
We selected three commonly used models to assess the effectiveness of various pretrained backbones: the text-only RoBERTa \cite{roberta}, the text and layout-aware LiLT \cite{lilt}, and the text, layout, and vision-aware LayoutLMv3 \cite{layoutlmv3}.
Table~\ref{tab:warmer_backbone} shows that multimodal frameworks tend to outperform the monomodal RoBERTa, particularly when OCR errors impact the input text sequence. However, LayoutLMv3-Chinese exhibits weaker feature representation, significantly underperforming compared to LiLT and RoBERTa, despite all three using the same xlm-RoBERTa-base checkpoints. Interestingly, there are instances where the monomodal RoBERTa outperforms multimodal backbones, indicating that multimodal architectures do not always guarantee superior performance or enhanced domain-specific knowledge extraction. 
%
\begin{table}[t]
    \centering
    \small
    \resizebox{0.7\linewidth}{!}{%
    \begin{tabular}{@{}l c c c c c@{}}
        \toprule
        Model & F-P & F-H & CORD & Ephoie & FUNSD \\
        \midrule
        Roberta & 64.18 & 23.85 & \textbf{70.40} & 31.57 & 59.44 \\
        LiLT & 63.82 & 30.89 & 67.87 & \textbf{31.97} & \textbf{60.94} \\
        LayoutLMv3 & \textbf{65.75} & \textbf{31.63} & 66.37 & 25.48 & 59.86 \\
        \bottomrule
    \end{tabular}%
    }
    \caption{Results under different Warmer backbones. }
    \label{tab:warmer_backbone}
\end{table}

\subsection{Recursive Inference Results}
\label{sec:recursive_inference}
Here, we assessed how effectively the zero-shot trained Warmer enhances MLLM inference and explored the impact of the recursive inferencing mechanism across various MLLMs. 

\paragraph{Performance on Various MLLMs} 
Table~\ref{tab:mllm_performance} presents the results of two high-performing open-source models (InternVL and QWenVL) and the best-performing proprietary model (Gemini). The result shows that the inclusion of Warmer outputs consistently improves performance across all models and datasets. This suggests that Warmer successfully captures task-specific patterns that generic MLLMs might overlook.
\begin{table}[t]
    \centering
    \resizebox{\linewidth}{!}{
    \begin{tabular}{@{}lcccccccc@{}}
        \toprule
        & \multicolumn{2}{c}{F-P} & \multicolumn{2}{c}{F-H} & \multicolumn{2}{c}{CORD} & \multicolumn{2}{c}{Ephoie} \\
        \cmidrule(rl){2-3} \cmidrule(rl){4-5} \cmidrule(rl){6-7} \cmidrule(rl){8-9}
        Model & Vani. & Ours & Vani. & Ours & Vani. & Ours & Vani. & Ours \\
        \midrule
        InternVL 
        & 66.56 & \up{68.09}  
        & 45.47 & \up{46.81}  
        & 66.84 & \up{68.8}  
        & 68.92 & \up{70.29}  \\

        QWenVL  
        & 78.05 & \down{77.27}  
        & 43.65 & \up{44.43}  
        & 77.86 & \up{78.44}  
        & 70.36 & \up{75.03}  \\

        Gemini  
        & 76.09 & \up{81.91}  
        & 66.86 & \up{68.02}  
        & 84.35 & \up{85.19}  
        & 81.82 & \up{82.15}  \\
        
        \bottomrule
    \end{tabular}
    }
    \caption{Comparison of Warmer to Generative Models.}
    \label{tab:mllm_performance}
\end{table}

\begin{figure}[t]
    \centering
    \tiny \includegraphics[width=\linewidth]{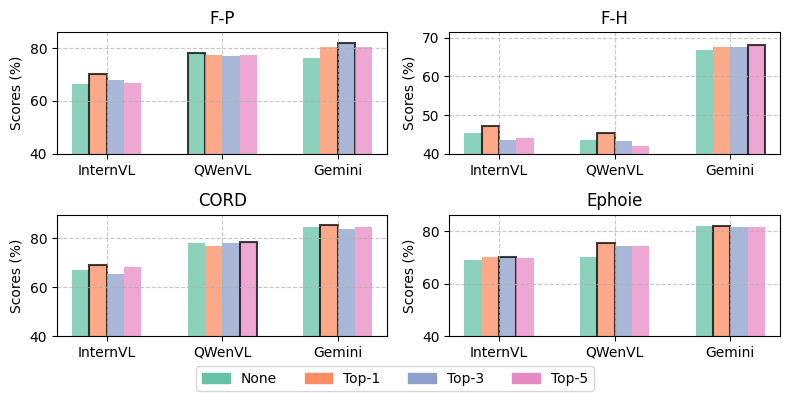}
    \caption{Result comparison by feeding Top-$K$ Warmer-Retrieved Candidates into MLLM.}
    \label{fig:top_k_outputs}
\end{figure}

\paragraph{Effectiveness of Top-$K$ Candidates.}
Figure~\ref{fig:top_k_outputs} shows that providing top-$K$ candidates from the warmer can enhance the likelihood of integrating relevant extracted information into MLLMs and improve performance. For instance, in FormNLU, retrieving additional information from the warmer can guide Gemini to focus on relevant context, thereby enhancing its performance. However, this approach also introduces the risk of incorporating noise into the prompt, which may negatively impact the generative model's performance. This effect is particularly notable in InternVL2 and QWenVL2, when applied to datasets with OCR-challenging like F-H and Ephoie. 

\paragraph{Effectiveness of Iterative Tuning.}
Recursive inference is introduced to enhance both Warmer retrieval and MLLM-generated answer quality. 
Table~\ref{tab:recursive_inference} shows that models exhibit improved performance when more than one iteration is conducted. This demonstrates that Warmer and the LLM generator can mutually reinforce each other, enabling the model to generate more accurate final predictions.
Additionally, we observed that open-source models (InternVL, QWenVL) typically require more iterations to reach peak performance, while the closed-source Gemini often achieves its best results with fewer iterations. Moreover, datasets that present OCR challenges (F-H and Ephoie) benefit from additional iterations, with all models requiring at least two iterations for optimal performance.

\definecolor{lightgreen}{RGB}{204, 255, 204} 

\begin{table}[t]
    \centering
    \renewcommand{\arraystretch}{1.05}
    \resizebox{\linewidth}{!}{ 
    \begin{tabular}{@{}lcccccccccccc@{}}
        \toprule
        Iter. & \multicolumn{3}{c}{F-P} & \multicolumn{3}{c}{F-H} & \multicolumn{3}{c}{CORD} & \multicolumn{3}{c}{EPHOIE} \\
        \cmidrule(lr){2-4} \cmidrule(lr){5-7} \cmidrule(lr){8-10} \cmidrule(l){11-13}
        & Int & Qw & Gemi & Int & QW & Gemi & Int & QW & Gemi & Int & QW & Gemi \\
        \midrule
        Vani.   & 66.56 & 78.05 & 76.09 & 45.47 & 43.65 & 66.86 & 66.84 & 77.86 & 84.35 & 68.92 & 70.36 & 81.82 \\
        Iter 1 & 68.09 & 76.53 & \cellcolor{lightgreen}\textbf{80.29} & 46.81 & 44.43 & 67.73 & \cellcolor{lightgreen}\textbf{68.80} & 76.93 & \cellcolor{lightgreen}\textbf{85.19} & 68.54 & 75.03 & 81.80 \\
        Iter 2 & \cellcolor{lightgreen}\textbf{70.12} & 77.22 & 80.17 & 46.17 & \cellcolor{lightgreen}\textbf{45.27} & 67.60 & 67.89 & 76.70 & 84.67 & 69.49 & \cellcolor{lightgreen} \textbf{75.55} & \cellcolor{lightgreen}\textbf{81.91} \\
        Iter 3 & 68.54 & 76.75 & 80.15 & \cellcolor{lightgreen}\textbf{47.23} & 44.50 & 67.32 & 67.29 & 76.93 & 84.65 & \cellcolor{lightgreen}\textbf{70.24} & 75.44 & 81.71 \\
        Iter 4 & 68.28 & \cellcolor{lightgreen}\textbf{77.27} & 79.88 & 45.54 & 45.26 & \cellcolor{lightgreen}\textbf{67.63} & 66.84 & 76.70 & 84.39 & 68.99 & 75.55 & 82.15 \\
        Iter 5 & 70.21 & 76.75 & 80.06 & 44.86 & 44.51 & 67.63 & 67.28 & \cellcolor{lightgreen}\textbf{76.93} & 84.40 & 70.07 & 75.44 & 81.86 \\
        \bottomrule
    \end{tabular}
    }
    \caption{Performance trends of iterative tuning.  Int: InternVL2; QW: QWenVL2; Gemi: Gemini.}
    \label{tab:recursive_inference}
\end{table}

\paragraph{Recursive Warmer Performance.}
Table~\ref{tab:iterative_warmer} shows that recursive inference enhances both discriminative Warmer and generative MLLM performance. Notably, the FormNLU dataset exhibits significant improvement, with scores rising from 66.19 to 73.76 on the printed set and from 31.64 to 39.15 on the handwritten set. An interesting finding is that the performance peaks for Warmer and MLLM do not always coincide at the same iteration. This may suggest that while Warmer improves retrieval, Gemini might not immediately capitalize on these improvements due to its integration and reasoning process.
\begin{table}[t]
    \centering
    \resizebox{\linewidth}{!}{%
    \begin{tabular}{lcccccccc}
        \toprule
        & \multicolumn{2}{c}{F-P} & \multicolumn{2}{c}{F-H} & \multicolumn{2}{c}{CORD} & \multicolumn{2}{c}{Ephoie} \\
        \cmidrule(rl){2-3}\cmidrule(rl){4-5}\cmidrule(rl){6-7}\cmidrule(rl){8-9}
        & Warmer & Gemini & Warmer & Gemini & Warmer & Gemini & Warmer & Gemini \\
        \midrule
        Vanilla & 66.19 & 76.09 & 31.64 & 66.86 & \cellcolor{blue!20} \textbf{67.26} & 84.35 & 27.16 & 81.82 \\
        1 & 73.57 & \cellcolor{green!20}\textbf{80.29} & 38.11 & \cellcolor{green!20}\textbf{67.73} & 63.37 & \cellcolor{green!20}\textbf{85.19} & \cellcolor{blue!20} \textbf{27.98} & 81.80 \\
        2 & \cellcolor{blue!20}\textbf{73.76} & 80.17 & 38.79 & 67.60 & 64.15 & 84.67 & 25.94 & 81.91 \\
        3 & 73.72 & 80.15 & \cellcolor{blue!20}\textbf{39.15} & 67.32 & 64.32 & 84.65 & 26.03 & 81.71 \\
        4 & 73.76 & 79.88 & 38.84 & 67.63 & 64.32 & 84.39 & 25.94 & \cellcolor{green!20}\textbf{82.15} \\
        5 & 73.60 & 80.06 & 38.92 & 67.63 & 64.04 & 84.40 & 26.12 & 81.86 \\
        \bottomrule
    \end{tabular}%
    }
    \caption{Impact of number of iterations on Warmer and Gemini.}
    \label{tab:iterative_warmer}
\end{table}

\begin{figure}[t]
    \centering
    \tiny \includegraphics[width=\linewidth]{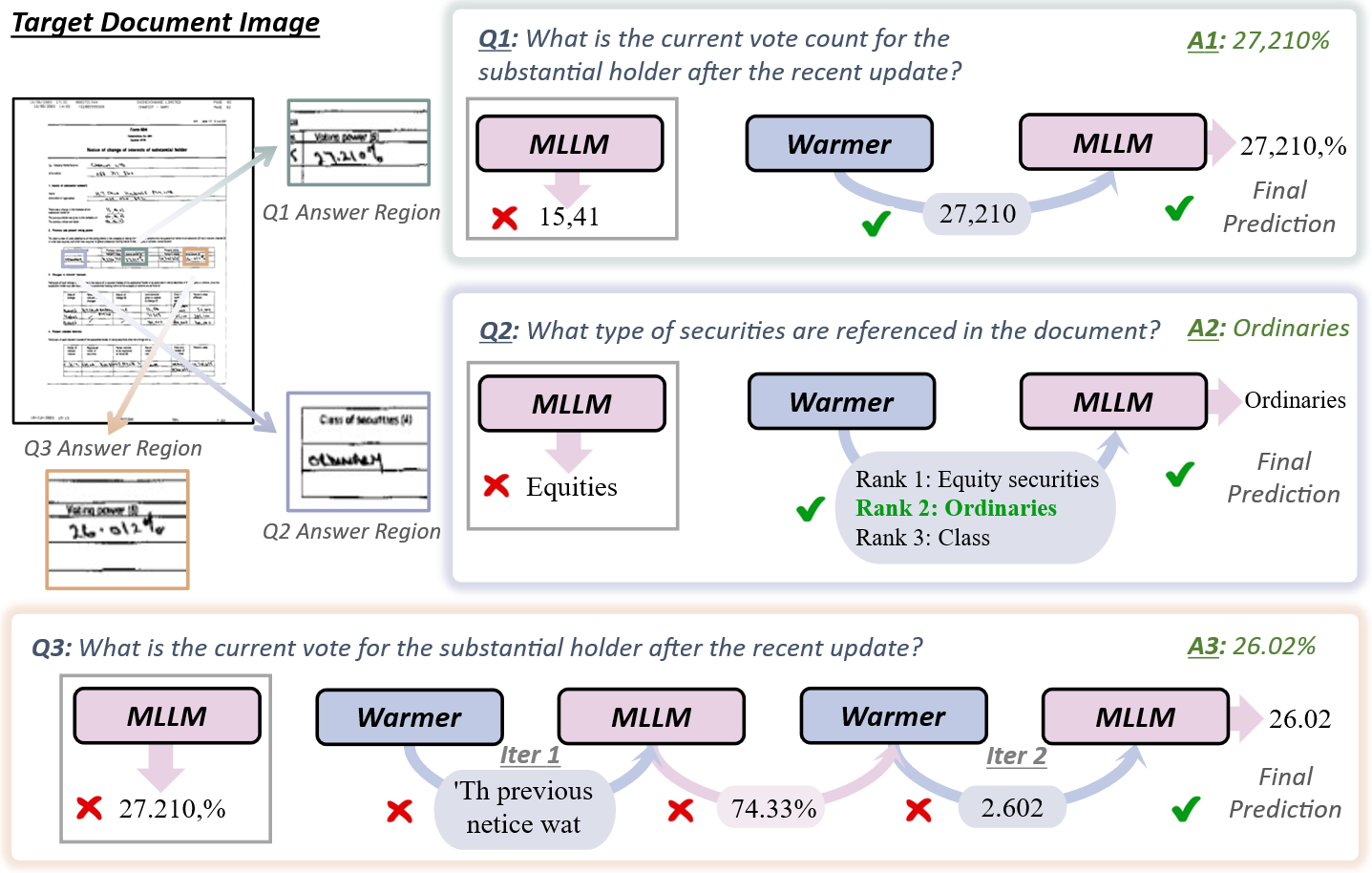}
    \caption{Qualitative Case Studies for Q1) Effectiveness of Warmer Retrieving; Q2) Demonstrating the Top-K candidates; 3) Effectiveness of iterative Inferencing.}
    \label{fig:case_study}
\end{figure}

\section{Case Study}
To further illustrate the effectiveness of SynDoc, Figure~\ref{fig:case_study} visualizes several examples where initial MLLM predictions are refined using SynDoc \footnote{Please refer to Appendix for more case  studies.}. In \textit{\textbf{Q1}}, a question regarding the present voting count initially yields an incorrect answer of 15,41, which is subsequently corrected to 27,210 with the aid of the warmer. This example highlights how the warmer effectively introduces domain-specific knowledge, mitigating hallucinations and reducing the imprecision of MLLM predictions.

Additionally, relying solely on the Top-1 retrieved answer from the warmer may not always capture the most relevant information needed for accurate answering. As demonstrated in \textit{\textbf{Q2}}, providing Top-3 entities enhances performance by leveraging both the warmer’s domain knowledge and the MLLM’s general world knowledge, thereby refining the final prediction.

The last example \textit{\textbf{Q3}} highlights the effectiveness of the iterative inference mechanism. Here, the warmer and MLLM incrementally improve each other's performance, leading to an almost correct prediction. Notably, even when the warmer provides the perfect hints in the final iteration, OCR errors may still be present. However, the MLLM compensates by leveraging its large-scale general world knowledge to generate the correct prediction.

\section{Conclusion}
In this paper, we introduced a novel VRDU framework, SynDoc, which effectively integrates discriminative VLPMs and generative MLLMs to advance domain-specific VRDU performance, particularly in zero-shot settings. Our extensive experiments show that the proposed \textit{Synthetic Data Generator} and \textit{Adaptive Warmer Tuning} enable the discriminative warmer to efficiently acquire domain knowledge and, together with recursive inference, drive continual performance gains for both the warmer and the MLLM. While the framework exhibits robust results on multiple domain-specific datasets, however, further enhancements may be required to maximize generalizability and robustness in cross-domain applications.






\bibliographystyle{elsarticle-num} 
\bibliography{custom}

\begin{thebibliography}{10}
\expandafter\ifx\csname url\endcsname\relax
  \def\url#1{\texttt{#1}}\fi
\expandafter\ifx\csname urlprefix\endcsname\relax\def\urlprefix{URL }\fi
\expandafter\ifx\csname href\endcsname\relax
  \def\href#1#2{#2} \def\path#1{#1}\fi

\bibitem{pdfvqa}
Y.~Ding, S.~Luo, H.~Chung, S.~C. Han, Pdf-vqa: A new dataset for real-world vqa on pdf documents, in: Machine Learning and Knowledge Discovery in Databases: Applied Data Science and Demo Track, Springer Nature Switzerland, 2023, pp. 585--601.

\bibitem{mmvqa}
Y.~Ding, K.~Ren, J.~Huang, S.~Luo, S.~C. Han, Mmvqa: A comprehensive dataset for investigating multipage multimodal information retrieval in pdf-based visual question answering, in: Proceedings of the Thirty-Third International Joint Conference on Artificial Intelligence, IJCAI, 2024, pp. 3--9.

\bibitem{formnlu}
Y.~Ding, S.~Long, J.~Huang, K.~Ren, X.~Luo, H.~Chung, S.~C. Han, Form-nlu: Dataset for the form natural language understanding, in: Proceedings of the 46th International ACM SIGIR Conference on Research and Development in Information Retrieval, 2023, pp. 2807--2816.

\bibitem{khalighinejad2024matvix}
G.~Khalighinejad, S.~Scott, O.~Liu, K.~L. Anderson, R.~Stureborg, A.~Tyagi, B.~Dhingra, Matvix: Multimodal information extraction from visually rich articles, arXiv preprint arXiv:2410.20494 (2024).

\bibitem{vrdu}
Z.~Wang, Y.~Zhou, W.~Wei, C.-Y. Lee, S.~Tata, Vrdu: A benchmark for visually-rich document understanding, in: Proceedings of the 29th ACM SIGKDD Conference on Knowledge Discovery and Data Mining, 2023, pp. 5184--5193.

\bibitem{layoutlmv3}
Y.~Huang, T.~Lv, L.~Cui, Y.~Lu, F.~Wei, Layoutlmv3: Pre-training for document ai with unified text and image masking, in: Proceedings of the 30th ACM International Conference on Multimedia, 2022, pp. 4083--4091.

\bibitem{udoc}
J.~Gu, J.~Kuen, V.~I. Morariu, H.~Zhao, R.~Jain, N.~Barmpalios, A.~Nenkova, T.~Sun, Unidoc: Unified pretraining framework for document understanding, Advances in Neural Information Processing Systems 34 (2021) 39--50.

\bibitem{lyu2024structextv3}
P.~Lyu, Y.~Li, H.~Zhou, W.~Ma, X.~Wan, Q.~Xie, L.~Wu, C.~Zhang, K.~Yao, E.~Ding, et~al., Structextv3: An efficient vision-language model for text-rich image perception, comprehension, and beyond, arXiv preprint arXiv:2405.21013 (2024).

\bibitem{david}
Y.~Ding, S.~C. Han, Z.~Li, H.~Chung, David: Domain adaptive visually-rich document understanding with synthetic insights, arXiv preprint arXiv:2410.01609 (2024).

\bibitem{gpt4o}
{OpenAI}, Hello gpt-4o, \url{https://openai.com/index/hello-gpt-4o/} (2024).

\bibitem{Qwen2-VL}
P.~Wang, S.~Bai, S.~Tan, S.~Wang, Z.~Fan, J.~Bai, K.~Chen, X.~Liu, J.~Wang, W.~Ge, Y.~Fan, K.~Dang, M.~Du, X.~Ren, R.~Men, D.~Liu, C.~Zhou, J.~Zhou, J.~Lin, Qwen2-vl: Enhancing vision-language model's perception of the world at any resolution, arXiv preprint arXiv:2409.12191 (2024).

\bibitem{hu2024mplug2}
A.~Hu, H.~Xu, L.~Zhang, J.~Ye, M.~Yan, J.~Zhang, Q.~Jin, F.~Huang, J.~Zhou, mplug-docowl2: High-resolution compressing for ocr-free multi-page document understanding, CoRR (2024).

\bibitem{feng2024docpedia}
H.~Feng, Q.~Liu, H.~Liu, J.~Tang, W.~Zhou, H.~Li, C.~Huang, Docpedia: Unleashing the power of large multimodal model in the frequency domain for versatile document understanding, Science China Information Sciences 67~(12) (2024) 1--14.

\bibitem{tang2024textsquare}
J.~Tang, C.~Lin, Z.~Zhao, S.~Wei, B.~Wu, Q.~Liu, H.~Feng, Y.~Li, S.~Wang, L.~Liao, et~al., Textsquare: Scaling up text-centric visual instruction tuning, arXiv preprint arXiv:2404.12803 (2024).

\bibitem{hu2024mplug}
A.~Hu, H.~Xu, J.~Ye, M.~Yan, L.~Zhang, B.~Zhang, J.~Zhang, Q.~Jin, F.~Huang, J.~Zhou, mplug-docowl 1.5: Unified structure learning for ocr-free document understanding, in: Findings of the Association for Computational Linguistics: EMNLP 2024, 2024, pp. 3096--3120.

\bibitem{watanabe1995layout}
T.~Watanabe, Q.~Luo, N.~Sugie, Layout recognition of multi-kinds of table-form documents, IEEE Transactions on Pattern Analysis and Machine Intelligence 17~(4) (1995) 432--445.

\bibitem{seki2007information}
M.~Seki, M.~Fujio, T.~Nagasaki, H.~Shinjo, K.~Marukawa, Information management system using structure analysis of paper/electronic documents and its applications, in: Ninth International Conference on Document Analysis and Recognition (ICDAR 2007), Vol.~2, IEEE, 2007, pp. 689--693.

\bibitem{oliveira2017fast}
D.~A.~B. Oliveira, M.~P. Viana, Fast cnn-based document layout analysis, in: 2017 IEEE International Conference on Computer Vision Workshops (ICCVW), IEEE, 2017, pp. 1173--1180.

\bibitem{udop}
Z.~Tang, Z.~Yang, G.~Wang, Y.~Fang, Y.~Liu, C.~Zhu, M.~Zeng, C.~Zhang, M.~Bansal, Unifying vision, text, and layout for universal document processing, in: Proceedings of the IEEE/CVF Conference on Computer Vision and Pattern Recognition, 2023, pp. 19254--19264.

\bibitem{lilt}
J.~Wang, L.~Jin, K.~Ding, Lilt: A simple yet effective language-independent layout transformer for structured document understanding, in: Proceedings of the 60th Annual Meeting of the Association for Computational Linguistics (Volume 1: Long Papers), 2022, pp. 7747--7757.

\bibitem{bros}
T.~Hong, D.~Kim, M.~Ji, W.~Hwang, D.~Nam, S.~Park, Bros: A pre-trained language model focusing on text and layout for better key information extraction from documents, in: Proceedings of the AAAI Conference on Artificial Intelligence, Vol.~36, 2022, pp. 10767--10775.

\bibitem{rvlcdip}
A.~W. Harley, A.~Ufkes, K.~G. Derpanis, Evaluation of deep convolutional nets for document image classification and retrieval, in: 2015 13th International Conference on Document Analysis and Recognition (ICDAR), IEEE, 2015, pp. 991--995.

\bibitem{funsd}
G.~Jaume, H.~K. Ekenel, J.-P. Thiran, Funsd: A dataset for form understanding in noisy scanned documents, in: 2019 International Conference on Document Analysis and Recognition Workshops (ICDARW), Vol.~2, IEEE, 2019, pp. 1--6.

\bibitem{cord}
S.~Park, S.~Shin, B.~Lee, J.~Lee, J.~Surh, M.~Seo, H.~Lee, Cord: a consolidated receipt dataset for post-ocr parsing, in: Workshop on Document Intelligence at NeurIPS 2019, 2019.

\bibitem{zhang2024llava}
R.~Zhang, Y.~Zhou, J.~Chen, J.~Gu, C.~Chen, T.~Sun, Llava-read: Enhancing reading ability of multimodal language models, arXiv preprint arXiv:2407.19185 (2024).

\bibitem{ernie-mmlayout}
W.~Wang, Z.~Huang, B.~Luo, Q.~Chen, Q.~Peng, Y.~Pan, W.~Yin, S.~Feng, Y.~Sun, D.~Yu, et~al., Ernie-mmlayout: Multi-grained multimodal transformer for document understanding, arXiv preprint arXiv:2209.08569 (2022).

\bibitem{docformerv2}
S.~Appalaraju, P.~Tang, Q.~Dong, N.~Sankaran, Y.~Zhou, R.~Manmatha, Docformerv2: Local features for document understanding, arXiv preprint arXiv:2306.01733 (2023).

\bibitem{donut}
G.~Kim, T.~Hong, M.~Yim, J.~Nam, J.~Park, J.~Yim, W.~Hwang, S.~Yun, D.~Han, S.~Park, Ocr-free document understanding transformer, in: Computer Vision--ECCV 2022: 17th European Conference, Tel Aviv, Israel, October 23--27, 2022, Proceedings, Part XXVIII, Springer, 2022, pp. 498--517.

\bibitem{layoutxlm}
Y.~Xu, T.~Lv, L.~Cui, G.~Wang, Y.~Lu, D.~Florencio, C.~Zhang, F.~Wei, Layoutxlm: Multimodal pre-training for multilingual visually-rich document understanding, arXiv preprint arXiv:2104.08836 (2021).

\bibitem{visfocus}
O.~Abramovich, N.~Nayman, S.~Fogel, I.~Lavi, R.~Litman, S.~Tsiper, R.~Tichauer, S.~Appalaraju, S.~Mazor, R.~Manmatha, Visfocus: Prompt-guided vision encoders for ocr-free dense document understanding, in: European Conference on Computer Vision, Springer, 2024, pp. 241--259.

\bibitem{gemini1.5}
G.~Team, P.~Georgiev, V.~I. Lei, R.~Burnell, L.~Bai, et~al., \href{https://arxiv.org/abs/2403.05530}{Gemini 1.5: Unlocking multimodal understanding across millions of tokens of context} (2024).
\newblock \href {http://arxiv.org/abs/2403.05530} {\path{arXiv:2403.05530}}.
\newline\urlprefix\url{https://arxiv.org/abs/2403.05530}

\bibitem{qwen}
J.~Bai, S.~Bai, S.~Yang, S.~Wang, S.~Tan, P.~Wang, J.~Lin, C.~Zhou, J.~Zhou, Qwen-vl: A versatile vision-language model for understanding, localization, text reading, and beyond (2023).

\bibitem{Idefics2}
H.~Laurençon, L.~Tronchon, M.~Cord, V.~Sanh, \href{https://arxiv.org/abs/2405.02246}{What matters when building vision-language models?} (2024).
\newblock \href {http://arxiv.org/abs/2405.02246} {\path{arXiv:2405.02246}}.
\newline\urlprefix\url{https://arxiv.org/abs/2405.02246}

\bibitem{icld3ie}
J.~He, L.~Wang, Y.~Hu, N.~Liu, H.~Liu, X.~Xu, H.~T. Shen, Icl-d3ie: In-context learning with diverse demonstrations updating for document information extraction, in: Proceedings of the IEEE/CVF International Conference on Computer Vision, 2023, pp. 19485--19494.

\bibitem{lxmert}
H.~Tan, M.~Bansal, Lxmert: Learning cross-modality encoder representations from transformers, in: Proceedings of the 2019 Conference on Empirical Methods in Natural Language Processing and the 9th International Joint Conference on Natural Language Processing (EMNLP-IJCNLP), 2019, pp. 5100--5111.

\bibitem{vies}
J.~Wang, C.~Liu, L.~Jin, G.~Tang, J.~Zhang, S.~Zhang, Q.~Wang, Y.~Wu, M.~Cai, Towards robust visual information extraction in real world: New dataset and novel solution, in: Proceedings of the AAAI Conference on Artificial Intelligence, Vol.~35, 2021, pp. 2738--2745.

\bibitem{luo2024layoutllm}
C.~Luo, Y.~Shen, Z.~Zhu, Q.~Zheng, Z.~Yu, C.~Yao, Layoutllm: Layout instruction tuning with large language models for document understanding, arXiv preprint arXiv:2404.05225 (2024).

\bibitem{docvqa}
M.~Mathew, D.~Karatzas, C.~Jawahar, Docvqa: A dataset for vqa on document images, in: Proceedings of the IEEE/CVF winter conference on applications of computer vision, 2021, pp. 2200--2209.

\bibitem{internvl}
Z.~Chen, J.~Wu, W.~Wang, W.~Su, G.~Chen, S.~Xing, M.~Zhong, Q.~Zhang, X.~Zhu, L.~Lu, et~al., Internvl: Scaling up vision foundation models and aligning for generic visual-linguistic tasks, in: Proceedings of the IEEE/CVF Conference on Computer Vision and Pattern Recognition, 2024, pp. 24185--24198.

\bibitem{roberta}
Y.~Liu, Roberta: A robustly optimized bert pretraining approach, arXiv preprint arXiv:1907.11692 364 (2019).

\end{thebibliography}




\appendix

\clearpage
\section{Detailed Dataset Information}
\subsection{Dataset Description}
\label{app:dataset_description}
\noindent\textbf{Form-NLU} \cite{formnlu} is introduced for financial-domain form layout and content understanding, focusing on single-template, multi-format forms, including digital, printed, and handwritten variations. This dataset specifically addresses KIE tasks, which involve extracting 12 types of key information from more challenging printed and handwritten documents. Examples of these key information fields include "\textit{Substantial Holder Name}", "\textit{Previous Persons' Votes}", and others.

\noindent\textbf{CORD} \cite{cord} is proposed for receipt understanding with diverse receipt templates. This dataset focuses on the sub-task of KIE to extract fine-grained key information from scanned receipts, such as "\textit{store name}" and "\textit{item quantity}".

\noindent\textbf{Ephoie} \cite{vies} is a dataset proposed for understanding scanned Chinese exam paper headers. The collected exam papers have diverse templates and handwritten information. This dataset focuses on the KIE sub-task to extract information from these exam papers, such as "\textit{Score}," "\textit{School}," and "\textit{Student Name}."

\noindent\textbf{FUNSD} \cite{funsd} is a dataset for form understanding, comprising scanned form images from diverse sources with varying templates. Each form contains predefined key-value pairs categorized as "\textit{Question}" and "\textit{Answer}" in the metadata. This dataset is utilized to assess the capability of the proposed framework in handling cross-domain scenarios.

\begin{table}[h]
    \centering
    \resizebox{\linewidth}{!}{%
    \renewcommand{\arraystretch}{1.2} 
    \begin{tabular}{l|l|c|c|c|c|c|c}
        \hline
        \textbf{Domain} & \textbf{Category} & \textbf{\# Doc} & \textbf{\# QA} & \textbf{Set 1} & \textbf{Set 2} & \textbf{Set 3} & \textbf{Set 4} \\
        \hline
        FormNLU-P & Financial Form & 50 & 596 & 1937 & 1137 & 1073 & 676 \\
        FormNLU-H & Financial Form & 50 & 597 & 1998 & 621 & 815 & 302 \\
        CORD & Receipt & 100 & 156 & 1644 & 1535 & 988 & 968 \\
        EPHOIE & Exam Paper & 311 & 928 & 2488 & 1746 & 1553 & 1159 \\
        FUNSD & Cross-domain & 50 & 467 & 2036 & 1905 & 1088 & 1022 \\
        \hline
    \end{tabular}}
    \caption{Dataset statistics across different dataset including the size of original test and the synthetic dataset. }
    \label{tab:dataset_stats}
\end{table}

\section{Detailed Model Information}
\label{app:model_details}

\subsection{Warmer Variants Details}

\noindent\textbf{RoBERTa} \cite{roberta}: 
RoBERTa is a self-supervised text-only language model trained on a large corpus, including BookCorpus, English Wikipedia, CommonCrawl News, OpenWebText, and Stories datasets. RoBERTa removes the next-sentence prediction (NSP) objective and uses dynamic masking, larger batch sizes, and longer sequences.

\noindent\textbf{LiLT} \cite{lilt}: 
LiLT (Language-independent Layout Transformer) extends pretrained text encoders with a lightweight layout encoder. It is pretrained on the IIT-CDIP scanned document corpus. LiLT features a dual-stream architecture to separately encode text and layout (bounding box) information, with Bi-directional Attention Complementation (BiACM) to enhance cross-modal alignment.

\noindent\textbf{LayoutLMv3} \cite{layoutlmv3}: 
LayoutLMv3 is a multimodal Transformer that jointly encodes text, layout, and image information. It is pretrained on the IIT-CDIP corpus and synthetic document data, using masked language modeling (MLM), masked image modeling (MIM), and word-patch alignment (WPA) tasks.

\subsection{Large Vision-Language Models details}

\subsubsection{Close Source Models}
\noindent\textbf{GPT-4o} \cite{gpt4o}: 
GPT-4o is a multimodal model capable of processing text, images, and audio, with an estimated size in the hundreds of billions to 1 trillion parameters. Trained on web-scale text, images, and audio, GPT-4o features native multimodal reasoning, multilingual support, and high-speed inference.

\noindent\textbf{Gemini 1.5} \cite{gemini1.5}: 
Gemini 1.5 Pro is a mid-size multimodal model with a Mixture-of-Experts (MoE) architecture, trained on a vast multimodal corpus  with a focus on long-context tasks up to 1 million tokens. 

\subsubsection{Open Source Models}

\noindent\textbf{InternVL2} \cite{internvl}: 
InternVL2 combines a vision Transformer and a language model. It is pretrained on 5M curated multimodal samples, including documents, forms, scientific charts, and medical images. 
InternVL2 ranges from 1B to 108B parameters, pretrained on curated multimodal data including documents, forms, scientific charts, and medical images. It achieves competitive results on specific document-centric tasks, such as DocVQA.

\noindent\textbf{QwenVL2} \cite{Qwen2-VL}: 
QwenVL2 is trained on 1.4T tokens, including image-text pairs, OCR data, video, and interleaved documents. With innovations like Naive Dynamic Resolution and Multimodal RoPE, QwenVL2 achieves competitive performance on multimodal benchmarks, establishing itself as a leading open-source option.

\noindent\textbf{Idefics2} \cite{Idefics2}: 
Idefics2 combines a Mistral-7B language model with a SigLIP vision encoder. Trained on interleaved web documents, captions, OCR data, and diagram-text mappings, it supports arbitrary sequences of text and images. Despite its smaller size, it achieves comparable performance to 30B+ models.

\begin{table}[t]
\centering
\resizebox{0.8\linewidth}{!}{%
\begin{tabular}{l|l|l|l|l}
\hline
\textbf{Model} & \textbf{Params} & \textbf{Modality} & \textbf{Training Data} & \textbf{Status} \\
\hline
RoBERTa & 125M & Text & Web, Books & Open \\
LiLT & 131M & Text+Layout & IIT-CDIP & Open \\
LayoutLMv3 & 133M& Text+Layout+Vision & IIT-CDIP & Open \\
\hline
GPT-4o & $\sim$200B & Text+Vision+Audio & Web+Images+Audio & Closed \\
Gemini 1.5 & 175B & Text+Vision+Audio & Web+Multimodal & Closed \\
InternVL2 & 8B & Text+Vision & Documents, Medical & Open \\
QwenVL2 & 72B & Text+Vision+Video & Web, OCR, Video & Open \\
Idefics2 & 8B & Text+Vision & Web, Documents & Open \\
\hline
\end{tabular}}
\caption{Baseline Models for Visual-rich Document Understanding (Appendix)}
\label{tab:baselines}
\end{table}

\section{Detailed Prompts}
We list all the prompts used in this paper for synthetic data generation in Table ~\ref{tab:dataset_generation_prompt} and MLLM zero-shot testing in Table~\ref{tab:inference_prompt_summary}. 
\begin{table*}[t]
    \centering
    \footnotesize
    \renewcommand{\arraystretch}{1.2}
    \begin{tabular}{p{3cm}|p{3cm}|p{6cm}}
        \hline
        \textbf{Module } & \textbf{Prompt Description} & \textbf{Prompt Template} \\
        \hline
        User-Input Verification & Checks whether the target information was entered by the user or is part of the form template. &  
        Based on the provided Context \texttt{\{\}} from the target form and the form image itself, check if the target information itself (do not consider the context) "\texttt{\{\}}" was entered by the form user (not part of the form template). Only output "Yes" if the \texttt{\{\}} is exactly provided by user not from the form template, do not consider context information. The response should follow the format below: "Response": "Yes/No" \\
        \hline
        Semantic Question Generation & Generates a short human-asked question where the answer exactly matches the target. &  
        Based on the above context \texttt{\{\}} and target document image, generate a human-asked SHORT question (output question only) of which answer is exactly same as "\texttt{\{\}}" \\
        \hline
        Answer Verification & Verifies whether the given target could be the expected answer to the given question. & Ignore the context information and domain knowledge (e.g. FAX NUMBER). Just consider whether '\texttt{\{\}}' could be the expected answer to the question '\texttt{\{\}}'. Output format: \texttt{\{'Response': 'Yes/No', 'Explanation': 'xxx'\}}. \\
        \hline
        Layout-Aware Question Reformulation & Reformulates a question into a short question about the location of the answer in the document. & Change the question \texttt{\{\}} to a very short question about finding the position of the answer from input document image. For example, where is the answer of xx located? \\
        \hline
        \hline
    \end{tabular}
    \caption{Synthetic Data Generator Prompt Example}
    \label{tab:dataset_generation_prompt}
\end{table*}

\begin{table*}[h]
    \centering
    \footnotesize
\renewcommand{\arraystretch}{1.2}
    \begin{tabular}{p{3cm}|p{3cm}|p{6cm}}
        \hline
        \textbf{Module } & \textbf{Prompt Description} & \textbf{Prompt Template} \\
        \hline
        Text-Image QA without Tips & Generates a response to a question based on an image and text context, without any additional Tips. & 
        Above is the context \texttt{\{\}}  of the target \texttt{\{\}}. 
        Please answer the question '\texttt{\{\}}' based on the context and image. 
        The output format must strictly follow: \newline
        Answer: xxx \\
        \hline
        Text-Image QA with One Tip & Generates a response to a question based on an image and text context, with a single Tip. & 
        The above is the context \texttt{\{\}}  of the target \texttt{\{\}}. 
        This is a Tip: '\texttt{\{\}}' (which may not be correct). 
        Please answer the question '\texttt{\{\}}' based on the context and image. 
        The output format must strictly follow: \newline
        Answer: xxx \\
        \hline
        Text-Image QA with Multiple Tips & Generates a response to a question based on an image and text context, with multiple ranked Tips. & 
        The above is the context \texttt{\{\}}  of the target \texttt{\{\}}. 
        These are the Tips (which may not be correct):  \newline
        Please answer the question '\texttt{\{\}}' based on the context and image. 
        The output format must strictly follow: \newline
        Answer: xxx \\
        \hline
        Text-Image QA with Bounding Boxes (No Tips) & Generates a response to a question based on an image, text context, and bounding box overlays, without any additional Tips. & 
        Above is the context \texttt{\{\}}  of the target \texttt{\{\}} document, \newline
        Please answer the question \texttt{\{\}}, \newline
        Based on the context and image, \newline
        The output format strictly follows: \newline
        Answer: xxx \\
        \hline
        Text-Image QA with Bounding Boxes (One Tip) & Generates a response to a question based on an image, text context, and bounding box overlays, with a single Tip. & 
        The above is the context \texttt{\{\}} of the target \texttt{\{\}} document. \newline
        This is a Tip: '\texttt{\{\}}' (which may not be correct). \newline
        Please answer the question \texttt{\{\}}, \newline
        Based on the context and image, \newline
        The output format strictly follows: \newline
        Answer: xxx \\
        \hline
        Text-Image QA with Bounding Boxes (Multiple Tips) & Generates a response to a question based on an image, text context, and bounding box overlays, with multiple ranked Tips. & 
        The above is the context \texttt{\{\}} of the target \texttt{\{\}} document. \newline
        These are Tips: '\texttt{\{\}}', (which may not be correct.) \newline
        Please answer the question \texttt{\{\}}, \newline
        Based on the context and images, \newline
        The output format strictly follows: \newline
        Answer: xxx \\
        \hline
    \end{tabular}
    \caption{Summary of Inference Prompt Functions and Their Templates}
    \label{tab:inference_prompt_summary}
\end{table*}

\section{Computational Cost}
Table~\ref{tab:training_cost} presents the training and inference resource consumption across five benchmark datasets with a consistent batch size of 16. The GPU memory usage remains within a reasonable range (approximately 25.5GB–28GB), demonstrating the framework's efficiency and scalability on standard hardware. The structural and semantic training times per epoch are well-balanced, typically ranging from 2 to 8 minutes, depending on dataset complexity. Notably, the inference time remains minimal—under 2.5 minutes for all datasets—highlighting the framework’s practical deployment potential. These results indicate that the proposed framework achieves a favorable trade-off between training cost and performance, making it suitable for both research and real-world applications.
\begin{table*}[htbp]
\centering
\begin{adjustbox}{max width=\textwidth}
\begin{tabular}{l|c|c|c|c|c}
\toprule
\textbf{Dataset} & \textbf{Batch Size} & \textbf{GPU Consumption} & \textbf{Structural Time (1 Epoch)} & \textbf{Semantic Time (1 Epoch)} & \textbf{Inference Time} \\
\midrule
FormNLU-P & 16 & 27983.4M & 00:03:46 & 00:03:08 & 00:01:10 \\
FormNLU-H & 16 & 25736.0M & 00:03:58 & 00:03:01 & 00:01:02 \\
CORD      & 16 & 26174.5M & 00:04:30 & 00:04:02 & 00:02:01 \\
EPHOIE    & 16 & 27993.1M & 00:06:01 & 00:03:12 & 00:01:14 \\
FUNSD     & 16 & 25566.2M & 00:08:10 & 00:02:01 & 00:00:59 \\
\bottomrule
\end{tabular}
\end{adjustbox}
\caption{Per-epoch GPU consumption and time cost across different datasets with a fixed batch size of 16. The reported times correspond to the most effective training configurations: 2 epochs for structural adaptation and 10 epochs for semantic adaptation.}

\label{tab:training_cost}
\end{table*}

\section{Additional Evaluation Results}
\subsection{Various Prompt Method Performance}
We present the results obtained using various prompting methods for baseline MLLMs and the Gemini-based SynDoc framework. The findings indicate that multimodal prompting, which integrates OCR-extracted textual context with document images, generally enhances performance. However, the OCR Challenging dataset exhibits difficulties in certain cases. For image-only prompting, some open-source models demonstrate relatively lower performance. Consequently, our SynDoc framework adopts the Image + Text context prompt as the primary approach for overall evaluation and ablation studies.

\begin{table}[h]
    \centering
    \resizebox{\linewidth}{!}{%
    \begin{tabular}{l l c c c c c}
        \toprule
        \textbf{Models} & \textbf{Prompt} & \textbf{Formnlu-P} & \textbf{Formnlu-H} & \textbf{CORD} & \textbf{Ephoie} & \textbf{Funsd} \\
        \midrule
        InternVL2 & Context-only & 59.65 & 7.16 & 44.00 & 54.39 & 53.48 \\
        Qwen2-VL &  & 72.12 & 10.04 & 65.20 & 61.59 & 68.87 \\
        Idefics2 &  & 28.52 & 3.33 & 4.33 & 8.90 & 21.98 \\
        GPT-4o &  & 71.64 & 1.45 & 69.88 & 59.78 & 68.71 \\
        Gemini &  & 70.88 & 5.91 & 71.53 & 59.94 & 68.21 \\
        \midrule
        InternVL2 & Image-only & 68.28 & 48.85 & 62.86 & 63.92 & 74.85 \\
        Qwen2-VL &  & 79.17 & 55.35 & 75.85 & 83.79 & 83.06 \\
        Idefics2 &  & 46.97 & 35.64 & 51.54 & 2.97 & 58.48 \\
        GPT-4o &  & 74.81 & 56.51 & 77.63 & 62.23 & 80.32 \\
        Gemini &  & 79.78 & 66.29 & 81.48 & 76.07 & 83.79 \\
        \midrule
        InternVL2 & Context + Image & 66.56 & 45.47 & 66.84 & 68.92 & 74.95 \\
        Qwen2-VL &  & 79.71 & 55.33 & 79.12 & 83.35 & 82.77 \\
        Idefics2 &  & 57.54 & 33.31 & 54.45 & 15.22 & 62.11 \\
        GPT-4o &  & 76.16 & 56.49 & 79.05 & 79.40 & 80.05 \\
        Gemini &  & 76.09 & 66.86 & 84.35 & 81.82 & 83.56 \\
        \midrule
        SynDoc & Context + Image & 81.91 & 68.02 & 85.19 & 82.15 & 83.02 \\
        SynDoc & Context + Image + bbox & 80.93 & 68.13 & 85.40 & 82.08 & 83.87 \\
        \bottomrule
    \end{tabular}%
    }
    \caption{Performance comparison of various models on different datasets.}
    \label{tab:model_performance}
\end{table}

\subsection{More Detailed Experimental Results}
We provide the detailed experimental results of different configurations for the MLLM inferencing, from Table~\ref{tab:iteration_performance1} to Table~\ref{tab:top5_ephoie}.

\begin{table*}[t]
    \centering
    \resizebox{\textwidth}{!}{%
    \begin{tabular}{l c c c c c c c c c c c c}
        \toprule
        \textbf{Model} & \multicolumn{2}{c}{\textbf{Baseline}} & \multicolumn{2}{c}{\textbf{Iteration 1}} & \multicolumn{2}{c}{\textbf{Iteration 2}} & \multicolumn{2}{c}{\textbf{Iteration 3}} & \multicolumn{2}{c}{\textbf{Iteration 4}} & \multicolumn{2}{c}{\textbf{Iteration 5}} \\
        \cmidrule(lr){2-3} \cmidrule(lr){4-5} \cmidrule(lr){6-7} \cmidrule(lr){8-9} \cmidrule(lr){10-11} \cmidrule(lr){12-13}
        & Warmer & LLM & Warmer & LLM & Warmer & LLM & Warmer & LLM & Warmer & LLM & Warmer & LLM \\
        \midrule
        InternVL & 67.26 & 66.84 & 63.38 & 68.80 & 57.74 & 67.89 & 58.50 & 67.29 & 59.70 & 66.84 & 57.82 & 67.28 \\
        QWenVL (2B) & 67.26 & 12.17 & 63.38 & 16.36 & 59.75 & 16.75 & 59.86 & 16.43 & 59.75 & 16.75 & 59.86 & 16.43 \\
        QWenVL (7B) & 67.26 & 77.86 & 63.38 & 76.93 & 59.89 & 76.70 & 59.64 & 76.93 & 59.89 & 76.70 & 59.64 & 76.93 \\
        QWenVL (72B) & 67.26 & 79.12 & 63.38 & 78.02 & 59.98 & 77.96 & 60.30 & 77.81 & 59.98 & 77.96 & 60.30 & 77.81 \\
        Gemini & 67.26 & 84.35 & 63.37 & 85.19 & 64.15 & 84.67 & 64.32 & 84.65 & 64.32 & 84.39 & 64.04 & 84.40 \\
        \bottomrule
    \end{tabular}%
    }
    \caption{Performance comparison across iterations for different models on the CORD dataset with Top-1 warmer retrieved entity.}
    \label{tab:iteration_performance1}
\end{table*}

\begin{table*}[h]
    \centering
    \resizebox{\textwidth}{!}{%
    \begin{tabular}{l c c c c c c c c c c c c}
        \toprule
        \textbf{Model} & \multicolumn{2}{c}{\textbf{Baseline}} & \multicolumn{2}{c}{\textbf{Iteration 1}} & \multicolumn{2}{c}{\textbf{Iteration 2}} & \multicolumn{2}{c}{\textbf{Iteration 3}} & \multicolumn{2}{c}{\textbf{Iteration 4}} & \multicolumn{2}{c}{\textbf{Iteration 5}} \\
        \cmidrule(lr){2-3} \cmidrule(lr){4-5} \cmidrule(lr){6-7} \cmidrule(lr){8-9} \cmidrule(lr){10-11} \cmidrule(lr){12-13}
        & Warmer & LLM & Warmer & LLM & Warmer & LLM & Warmer & LLM & Warmer & LLM & Warmer & LLM \\
        \midrule
        InternVL & 66.19 & 66.56 & 73.57 & 68.09 & 68.65 & 70.12 & 70.32 & 68.54 & 69.20 & 68.28 & 69.19 & 70.21 \\

        QWenVL (2B) & 66.19 & 44.85 & 73.57 & 50.34 & 61.63 & 50.45 & 61.82 & 50.52 & 61.76 & 50.48 & 61.84 & 50.54 \\
        QWenVL (7B) & 66.19 & 78.05 & 73.57 & 76.53 & 72.52 & 77.22 & 73.18 & 76.75 & 72.61 & 77.27 & 73.18 & 76.75 \\
        QWenVL (72B) & 66.19 & 79.71 & 73.57 & 81.21 & 74.41 & 81.42 & 74.54 & 81.20 & 74.58 & 81.42 & 74.54 & 81.20 \\
        Gemini & 66.19 & 76.09 & 73.57 & 80.29 & 73.76 & 80.17 & 73.72 & 80.15 & 73.76 & 79.88 & 73.60 & 80.06 \\
        \bottomrule
    \end{tabular}%
    }
    \caption{Performance comparison across iterations for different models on the Printed dataset with Top-1 warmer retrieved entity.}
    \label{tab:printed_performance2}
\end{table*}

\begin{table*}[h]
    \centering
    \resizebox{\textwidth}{!}{%
    \begin{tabular}{l c c c c c c c c c c c c}
        \toprule
        \textbf{Model} & \multicolumn{2}{c}{\textbf{Baseline}} & \multicolumn{2}{c}{\textbf{Iteration 1}} & \multicolumn{2}{c}{\textbf{Iteration 2}} & \multicolumn{2}{c}{\textbf{Iteration 3}} & \multicolumn{2}{c}{\textbf{Iteration 4}} & \multicolumn{2}{c}{\textbf{Iteration 5}} \\
        \cmidrule(lr){2-3} \cmidrule(lr){4-5} \cmidrule(lr){6-7} \cmidrule(lr){8-9} \cmidrule(lr){10-11} \cmidrule(lr){12-13}
        & Warmer & LLM & Warmer & LLM & Warmer & LLM & Warmer & LLM & Warmer & LLM & Warmer & LLM \\
        \midrule
        InternVL & 31.64 & 45.47 & 38.11 & 46.81 & 32.29 & 46.17 & 32.70 & 47.23 & 32.06 & 45.54 & 32.76 & 44.86 \\
        QWenVL (2B) & 31.64 & 14.56 & 38.11 & 19.21 & 24.95 & 19.33 & 25.45 & 19.19 & 25.02 & 19.36 & 25.44 & 19.20 \\
        QWenVL (7B) & 31.64 & 43.65 & 38.11 & 44.43 & 34.51 & 45.27 & 35.25 & 44.50 & 34.83 & 45.26 & 35.26 & 44.51 \\
        QWenVL (72B) & 31.64 & 55.33 & 38.11 & 58.33 & 38.37 & 58.40 & 38.33 & 58.37 & 38.48 & 58.58 & 38.36 & 58.37 \\
        Gemini & 31.64 & 66.86 & 38.11 & 67.73 & 38.79 & 67.60 & 39.15 & 67.32 & 38.84 & 67.63 & 38.92 & 67.63 \\
        \bottomrule
    \end{tabular}%
    }
    \caption{Performance comparison across iterations for different models on the Handwritten dataset with Top-1 warmer retrieved entity.}
    \label{tab:handwritten_performance3}
\end{table*}

\begin{table*}[h]
    \centering
    \resizebox{\textwidth}{!}{%
    \begin{tabular}{l c c c c c c c c c c c c}
        \toprule
        \textbf{Model} & \multicolumn{2}{c}{\textbf{Baseline}} & \multicolumn{2}{c}{\textbf{Iteration 1}} & \multicolumn{2}{c}{\textbf{Iteration 2}} & \multicolumn{2}{c}{\textbf{Iteration 3}} & \multicolumn{2}{c}{\textbf{Iteration 4}} & \multicolumn{2}{c}{\textbf{Iteration 5}} \\
        \cmidrule(lr){2-3} \cmidrule(lr){4-5} \cmidrule(lr){6-7} \cmidrule(lr){8-9} \cmidrule(lr){10-11} \cmidrule(lr){12-13}
        & Warmer & LLM & Warmer & LLM & Warmer & LLM & Warmer & LLM & Warmer & LLM & Warmer & LLM \\
        \midrule
        InternVL & 27.16 & 68.92 & 27.98 & 68.54 & 25.78 & 69.49 & 25.94 & 70.24 & 26.00 & 68.99 & 25.96 & 70.07 \\
        QWenVL (2B) & 27.16 & 46.13 & 27.98 & 36.51 & 27.10 & 37.00 & 26.78 & 36.39 & 27.10 & 36.97 & 26.78 & 36.39 \\
        QWenVL (7B) & 27.16 & 70.36 & 27.98 & 75.03 & 26.79 & 75.55 & 26.76 & 75.44 & 26.79 & 75.55 & 26.76 & 75.44 \\
                QWenVL (72B) & 27.16 & 83.35 & 27.98 & 81.95 & 26.38 & 82.08 & 26.51 & 82.06 & 26.38 & 82.08 & 26.51 & 82.06 \\
        Gemini & 27.16 & 81.82 & 27.98 & 81.80 & 25.94 & 81.91 & 26.03 & 81.71 & 25.94 & 82.15 & 26.12 & 81.86 \\
        \bottomrule
    \end{tabular}%
    }
    \caption{Performance comparison across iterations for different models on the Ephoie dataset with Top-1 warmer retrieved entity.}
    \label{tab:ephoie_performance4}
\end{table*}

\begin{table*}[h]
    \centering
    \resizebox{\textwidth}{!}{%
    \begin{tabular}{l c c c c c c c c c c c c}
        \toprule
        \textbf{Model} & \multicolumn{2}{c}{\textbf{Baseline}} & \multicolumn{2}{c}{\textbf{Iteration 1}} & \multicolumn{2}{c}{\textbf{Iteration 2}} & \multicolumn{2}{c}{\textbf{Iteration 3}} & \multicolumn{2}{c}{\textbf{Iteration 4}} & \multicolumn{2}{c}{\textbf{Iteration 5}} \\
        \cmidrule(lr){2-3} \cmidrule(lr){4-5} \cmidrule(lr){6-7} \cmidrule(lr){8-9} \cmidrule(lr){10-11} \cmidrule(lr){12-13}
        & Warmer & LLM & Warmer & LLM & Warmer & LLM & Warmer & LLM & Warmer & LLM & Warmer & LLM \\
        \midrule
        InternVL & 61.24 & 74.95 & 59.64 & 73.18 & 58.30 & 72.13 & 58.44 & 73.41 & 58.97 & 73.57 & 58.98 & 73.12 \\
        QWenVL & 61.24 & 79.12 & 61.94 & 74.84 & 60.03 & 75.73 & 60.92 & 74.57 & 59.93 & 75.73 & 60.92 & 74.57 \\
        Gemini & 61.24 & 83.56 & 59.17 & 82.77 & 59.77 & 83.02 & 60.06 & 82.38 & 59.54 & 82.91 & 60.11 & 82.36 \\
        \bottomrule
    \end{tabular}%
    }
    \caption{Performance comparison across iterations for different models on the FUNSD dataset  with Top-1 warmer retrieved entity.}
    \label{tab:funsd_performance5}
\end{table*}

\begin{table*}[h]
    \centering
    \resizebox{\textwidth}{!}{%
    \begin{tabular}{l c c c c c c c c c c c c}
        \toprule
        \textbf{Model} & \multicolumn{2}{c}{\textbf{Baseline}} & \multicolumn{2}{c}{\textbf{Iteration 1}} & \multicolumn{2}{c}{\textbf{Iteration 2}} & \multicolumn{2}{c}{\textbf{Iteration 3}} & \multicolumn{2}{c}{\textbf{Iteration 4}} & \multicolumn{2}{c}{\textbf{Iteration 5}} \\
        \cmidrule(lr){2-3} \cmidrule(lr){4-5} \cmidrule(lr){6-7} \cmidrule(lr){8-9} \cmidrule(lr){10-11} \cmidrule(lr){12-13}
        & Warmer & LLM & Warmer & LLM & Warmer & LLM & Warmer & LLM & Warmer & LLM & Warmer & LLM \\
        \midrule
        InternVL & 67.26 & 66.84 & 63.38 & 61.61 & 60.19 & 65.31 & 53.73 & 64.75 & 53.52 & 61.70 & 54.06 & 62.22 \\
        QWenVL & 67.26 & 77.86 & 63.38 & 78.16 & 59.65 & 77.96 & 59.34 & 78.12 & 59.65 & 77.96 & 59.34 & 78.12 \\
        Gemini & 67.26 & 84.35 & 63.38 & 83.46 & 63.79 & 82.34 & 63.42 & 83.07 & 63.42 & 83.07 & 63.69 & 83.00 \\
        \bottomrule
    \end{tabular}%
    }
    \caption{Top-3 Performance comparison across iterations for different models on the CORD dataset.}
    \label{tab:top3_cord}
\end{table*}

\begin{table*}[h]
    \centering
    \resizebox{\textwidth}{!}{%
    \begin{tabular}{l c c c c c c c c c c c c}
        \toprule
        \textbf{Model} & \multicolumn{2}{c}{\textbf{Baseline}} & \multicolumn{2}{c}{\textbf{Iteration 1}} & \multicolumn{2}{c}{\textbf{Iteration 2}} & \multicolumn{2}{c}{\textbf{Iteration 3}} & \multicolumn{2}{c}{\textbf{Iteration 4}} & \multicolumn{2}{c}{\textbf{Iteration 5}} \\
        \cmidrule(lr){2-3} \cmidrule(lr){4-5} \cmidrule(lr){6-7} \cmidrule(lr){8-9} \cmidrule(lr){10-11} \cmidrule(lr){12-13}
        & Warmer & LLM & Warmer & LLM & Warmer & LLM & Warmer & LLM & Warmer & LLM & Warmer & LLM \\
        \midrule
        InternVL & 66.19 & 66.56 & 73.56 & 65.91 & 67.70 & 67.85 & 67.55 & 67.12 & 68.53 & 66.21 & 66.92 & 66.38 \\
        QWenVL & 66.19 & 78.05 & 73.57 & 77.08 & 72.93 & 76.60 & 72.81 & 76.63 & 72.53 & 76.72 & 72.80 & 76.67 \\
        Gemini & 66.19 & 76.09 & 73.99 & 81.60 & 74.12 & 81.91 & 74.30 & 81.63 & 74.01 & 81.58 & 74.28 & 81.46 \\
        \bottomrule
    \end{tabular}%
    }
    \caption{Top-3 Performance comparison across iterations for different models on the Printed dataset.}
    \label{tab:top3_printed}
\end{table*}

\begin{table*}[h]
    \centering
    \resizebox{\textwidth}{!}{%
    \begin{tabular}{l c c c c c c c c c c c c}
        \toprule
        \textbf{Model} & \multicolumn{2}{c}{\textbf{Baseline}} & \multicolumn{2}{c}{\textbf{Iteration 1}} & \multicolumn{2}{c}{\textbf{Iteration 2}} & \multicolumn{2}{c}{\textbf{Iteration 3}} & \multicolumn{2}{c}{\textbf{Iteration 4}} & \multicolumn{2}{c}{\textbf{Iteration 5}} \\
        \cmidrule(lr){2-3} \cmidrule(lr){4-5} \cmidrule(lr){6-7} \cmidrule(lr){8-9} \cmidrule(lr){10-11} \cmidrule(lr){12-13}
        & Warmer & LLM & Warmer & LLM & Warmer & LLM & Warmer & LLM & Warmer & LLM & Warmer & LLM \\
        \midrule
        InternVL & 31.64 & 45.47 & 38.11 & 43.48 & 32.64 & 43.24 & 30.92 & 42.02 & 31.75 & 43.15 & 31.93 & 43.52 \\
        QWenVL & 31.64 & 43.65 & 38.11 & 42.03 & 33.65 & 43.37 & 33.68 & 41.66 & 32.60 & 42.62 & 33.28 & 41.55 \\
        Gemini & 31.64 & 66.86 & 38.11 & 66.82 & 39.35 & 67.68 & 39.48 & 67.12 & 39.15 & 66.80 & 38.79 & 67.49 \\
        \bottomrule
    \end{tabular}%
    }
    \caption{Top-3 Performance comparison across iterations for different models on the Handwritten dataset.}
    \label{tab:top3_handwritten}
\end{table*}

\begin{table*}[h]
    \centering
    \resizebox{\textwidth}{!}{%
    \begin{tabular}{l c c c c c c c c c c c c}
        \toprule
        \textbf{Model} & \multicolumn{2}{c}{\textbf{Baseline}} & \multicolumn{2}{c}{\textbf{Iteration 1}} & \multicolumn{2}{c}{\textbf{Iteration 2}} & \multicolumn{2}{c}{\textbf{Iteration 3}} & \multicolumn{2}{c}{\textbf{Iteration 4}} & \multicolumn{2}{c}{\textbf{Iteration 5}} \\
        \cmidrule(lr){2-3} \cmidrule(lr){4-5} \cmidrule(lr){6-7} \cmidrule(lr){8-9} \cmidrule(lr){10-11} \cmidrule(lr){12-13}
        & Warmer & LLM & Warmer & LLM & Warmer & LLM & Warmer & LLM & Warmer & LLM & Warmer & LLM \\
        \midrule
        InternVL & 27.16 & 68.92 & 27.98 & 70.29 & 26.00 & 69.04 & 26.08 & 69.35 & 26.17 & 68.15 & 26.30 & 69.41 \\
        QWenVL & 27.16 & 70.36 & 27.98 & 73.91 & 26.36 & 74.29 & 26.68 & 74.18 & 26.28 & 74.29 & 26.68 & 74.18 \\
        Gemini & 27.16 & 81.82 & 27.98 & 81.18 & 26.23 & 81.13 & 26.25 & 81.16 & 26.27 & 81.43 & 26.10 & 81.32 \\
        \bottomrule
    \end{tabular}%
    }
    \caption{Top-3 Warmer Retrieved Entity Performance comparison across iterations for different models on the Ephoie dataset.}
    \label{tab:top3_ephoie}
\end{table*}

\begin{table*}[h]
    \centering
    \resizebox{\textwidth}{!}{%
    \begin{tabular}{l c c c c c c c c c c c c}
        \toprule
        \textbf{Model} & \multicolumn{2}{c}{\textbf{Baseline}} & \multicolumn{2}{c}{\textbf{Iteration 1}} & \multicolumn{2}{c}{\textbf{Iteration 2}} & \multicolumn{2}{c}{\textbf{Iteration 3}} & \multicolumn{2}{c}{\textbf{Iteration 4}} & \multicolumn{2}{c}{\textbf{Iteration 5}} \\
        \cmidrule(lr){2-3} \cmidrule(lr){4-5} \cmidrule(lr){6-7} \cmidrule(lr){8-9} \cmidrule(lr){10-11} \cmidrule(lr){12-13}
        & Warmer & LLM & Warmer & LLM & Warmer & LLM & Warmer & LLM & Warmer & LLM & Warmer & LLM \\
        \midrule
        InternVL & 67.26 & 66.84 & 63.37 & 64.25 & 57.02 & 63.21 & 54.63 & 68.18 & 55.93 & 66.76 & 57.87 & 65.13 \\
        QWenVL & 67.26 & 77.86 & 63.38 & 78.20 & 59.54 & 77.49 & 58.91 & 78.44 & 60.08 & 77.53 & 58.91 & 78.16 \\
        Gemini & 67.26 & 84.35 & 63.38 & 84.57 & 63.79 & 82.85 & 63.99 & 83.37 & 63.79 & 82.77 & 63.79 & 83.39 \\
        \bottomrule
    \end{tabular}%
    }
    \caption{Top-5 Performance comparison across iterations for different models on the CORD dataset.}
    \label{tab:top5_cord}
\end{table*}

\begin{table*}[h]
    \centering
    \resizebox{\textwidth}{!}{%
    \begin{tabular}{l c c c c c c c c c c c c}
        \toprule
        \textbf{Model} & \multicolumn{2}{c}{\textbf{Baseline}} & \multicolumn{2}{c}{\textbf{Iteration 1}} & \multicolumn{2}{c}{\textbf{Iteration 2}} & \multicolumn{2}{c}{\textbf{Iteration 3}} & \multicolumn{2}{c}{\textbf{Iteration 4}} & \multicolumn{2}{c}{\textbf{Iteration 5}} \\
        \cmidrule(lr){2-3} \cmidrule(lr){4-5} \cmidrule(lr){6-7} \cmidrule(lr){8-9} \cmidrule(lr){10-11} \cmidrule(lr){12-13}
        & Warmer & LLM & Warmer & LLM & Warmer & LLM & Warmer & LLM & Warmer & LLM & Warmer & LLM \\
        \midrule
        InternVL & 66.19 & 66.56 & 73.57 & 66.88 & 69.19 & 66.17 & 67.46 & 65.23 & 65.67 & 65.67 & 66.27 & 66.27 \\
        QWenVL & 66.19 & 78.05 & 73.57 & 76.35 & 72.22 & 77.01 & 72.66 & 76.67 & 72.34 & 77.27 & 72.70 & 76.21 \\
        Gemini & 66.19 & 76.09 & 73.58 & 80.10 & 73.35 & 80.35 & 73.70 & 80.20 & 73.40 & 80.36 & 73.54 & 80.08 \\
        \bottomrule
    \end{tabular}%
    }
    \caption{Top-5 Performance comparison across iterations for different models on the Printed dataset.}
    \label{tab:top5_printed}
\end{table*}

\begin{table*}[h]
    \centering
    \resizebox{\textwidth}{!}{%
    \begin{tabular}{l c c c c c c c c c c c c}
        \toprule
        \textbf{Model} & \multicolumn{2}{c}{\textbf{Baseline}} & \multicolumn{2}{c}{\textbf{Iteration 1}} & \multicolumn{2}{c}{\textbf{Iteration 2}} & \multicolumn{2}{c}{\textbf{Iteration 3}} & \multicolumn{2}{c}{\textbf{Iteration 4}} & \multicolumn{2}{c}{\textbf{Iteration 5}} \\
        \cmidrule(lr){2-3} \cmidrule(lr){4-5} \cmidrule(lr){6-7} \cmidrule(lr){8-9} \cmidrule(lr){10-11} \cmidrule(lr){12-13}
        & Warmer & LLM & Warmer & LLM & Warmer & LLM & Warmer & LLM & Warmer & LLM & Warmer & LLM \\
        \midrule
        InternVL & 31.64 & 45.47 & 38.11 & 43.82 & 32.79 & 44.13 & 33.66 & 43.78 & 31.87 & 41.55 & 32.11 & 43.22 \\
        QWenVL & 31.64 & 43.65 & 38.11 & 40.12 & 32.51 & 41.97 & 33.13 & 40.18 & 32.26 & 41.75 & 32.99 & 39.78 \\
        Gemini & 31.64 & 66.86 & 38.11 & 66.90 & 39.05 & 67.33 & 39.06 & 67.51 & 38.99 & 67.01 & 39.11 & 68.02 \\
        \bottomrule
    \end{tabular}%
    }
    \caption{Top-5 Performance comparison across iterations for different models on the Handwritten dataset.}
    \label{tab:top5_handwritten}
\end{table*}

\begin{table*}[h]
    \centering
    \resizebox{\textwidth}{!}{%
    \begin{tabular}{l c c c c c c c c c c c c}
        \toprule
        \textbf{Model} & \multicolumn{2}{c}{\textbf{Baseline}} & \multicolumn{2}{c}{\textbf{Iteration 1}} & \multicolumn{2}{c}{\textbf{Iteration 2}} & \multicolumn{2}{c}{\textbf{Iteration 3}} & \multicolumn{2}{c}{\textbf{Iteration 4}} & \multicolumn{2}{c}{\textbf{Iteration 5}} \\
        \cmidrule(lr){2-3} \cmidrule(lr){4-5} \cmidrule(lr){6-7} \cmidrule(lr){8-9} \cmidrule(lr){10-11} \cmidrule(lr){12-13}
        & Warmer & LLM & Warmer & LLM & Warmer & LLM & Warmer & LLM & Warmer & LLM & Warmer & LLM \\
        \midrule
        InternVL & 27.16 & 68.92 & 27.98 & 68.88 & 26.18 & 68.66 & 25.93 & 67.90 & 26.06 & 69.61 & 25.87 & 69.77 \\
        QWenVL & 27.16 & 70.36 & 27.98 & 74.32 & 26.32 & 74.32 & 26.56 & 74.35 & 26.32 & 74.34 & 26.67 & 74.24 \\
        Gemini & 27.16 & 81.82 & 27.98 & 81.33 & 26.35 & 81.18 & 26.31 & 81.58 & 26.37 & 81.23 & 26.28 & 81.45 \\
        \bottomrule
    \end{tabular}%
    }
    \caption{Top-5 Performance comparison across iterations for different models on the Ephoie dataset.}
    \label{tab:top5_ephoie}
\end{table*}
\begin{figure}[t]
    \centering
    \tiny \includegraphics[width=\linewidth]{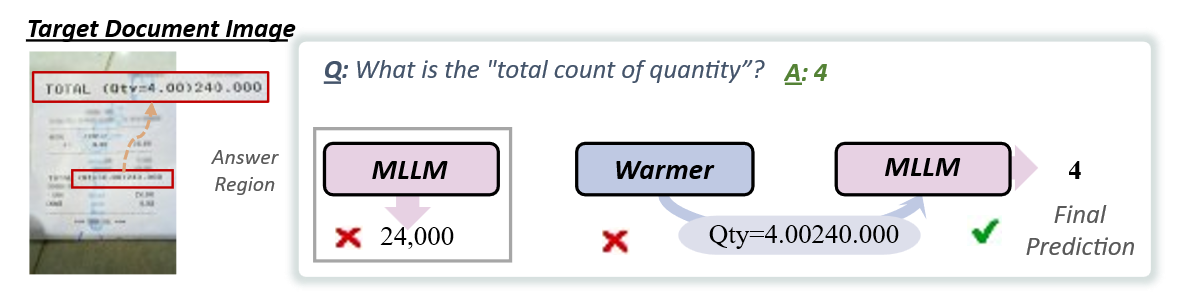}
    \caption{Qualitative case studies about CORD dataset for demonstrating the effectiveness of Warmer retrieved the content and the MLLM self-correction ability for OCR-error.}
    \label{fig:cord}
\end{figure}  
\begin{figure}[t]
    \centering
    \tiny \includegraphics[width=\linewidth]{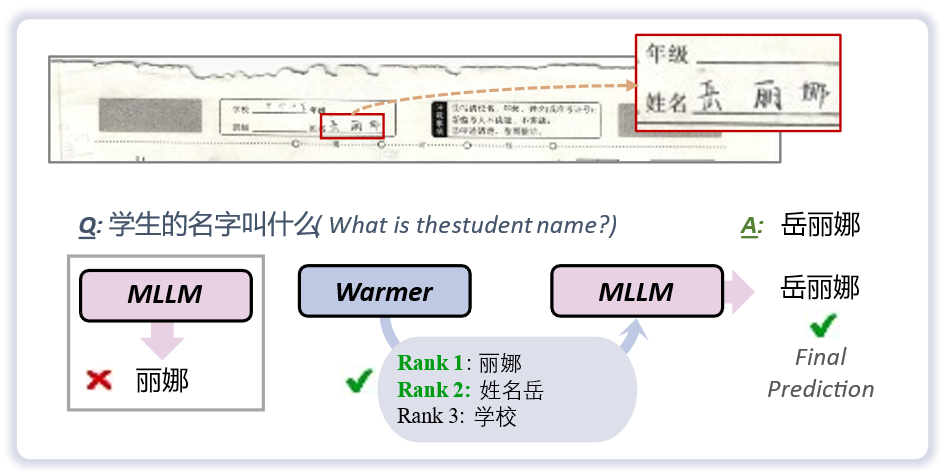}
    \caption{Qualitative case studies about Ephoie dataset for demonstrating the effectiveness of Top-$K$.}
    \label{fig:ephoie}
\end{figure}  
\section{Additionaly Case Studies}

Figures~\ref{fig:cord} and~\ref{fig:ephoie} present qualitative case studies from the CORD and Ephoie datasets, respectively, highlighting the  complementary strengths of MLLM-based self-correction pipeline and the Top-K retrieval. In Figure~\ref{fig:cord}, the MLLM initially predicts "24,000", and the Warmer module retrieves a noisy string "Qty=4.00240.000". Despite the noise, the final MLLM module successfully interprets the correct answer as "4", demonstrating its robustness to OCR errors and its ability to reason over imperfect retrieved content. In Figure~\ref{fig:ephoie}, a query about a student’s name is given, where the initial MLLM output is incorrect. However, the Warmer module retrieves relevant entities, ranking the correct answer within the Top-3, which enables the final MLLM stage to recover the accurate result. These examples collectively demonstrate the pipeline’s effectiveness in overcoming early-stage retrieval errors and OCR-related noise in complex document QA tasks.
\end{document}